\documentclass{article}

    \PassOptionsToPackage{numbers, compress}{natbib}
\usepackage[pdftex]{graphicx}
\usepackage{amsmath}
\usepackage[table,xcdraw]{xcolor}
\usepackage{graphicx}
\usepackage{subcaption}
\usepackage{tikz}
\usetikzlibrary{spy} 
\usepackage{dsfont}
\usepackage{colortbl}

\usepackage[inline]{enumitem}
\usepackage{algorithm}
\usepackage{algpseudocode}
\usepackage{amssymb} 
\usepackage{multirow}
\usepackage{pifont}

\newcommand{\tref}[1]{Table~\ref{#1}}
\newcommand{\fref}[1]{Figure~\ref{#1}}
\newcommand{\sref}[1]{Section~\ref{#1}}
\newcommand{\aref}[1]{Appendix~\ref{#1}}

\definecolor{third}{rgb}{1.0, 1.0, 0.6}
\definecolor{second}{rgb}{1.0, 0.85, 0.6}
\definecolor{first}{rgb}{1.0, 0.6, 0.6}

\usepackage[main, final]{neurips_2026}

\usepackage[utf8]{inputenc} 
\usepackage[T1]{fontenc}    
\usepackage{hyperref}       
\usepackage{url}            
\usepackage{booktabs}       
\usepackage{amsfonts}       
\usepackage{nicefrac}       
\usepackage{microtype}      
\usepackage{xcolor}         

\title{Warp-2DGS: Learnable Warped Kernels\\for 2D Gaussian Splatting}

\author{%
Huibin Li\quad Chul Min Yeum\\
CVISS Lab\\
University of Waterloo\\
{\tt\small \{h844li, cmyeum\}@uwaterloo.ca}
}

\begin{document}

\maketitle

\begin{figure}[htbp]
  \centering
  \includegraphics[width=1\textwidth]{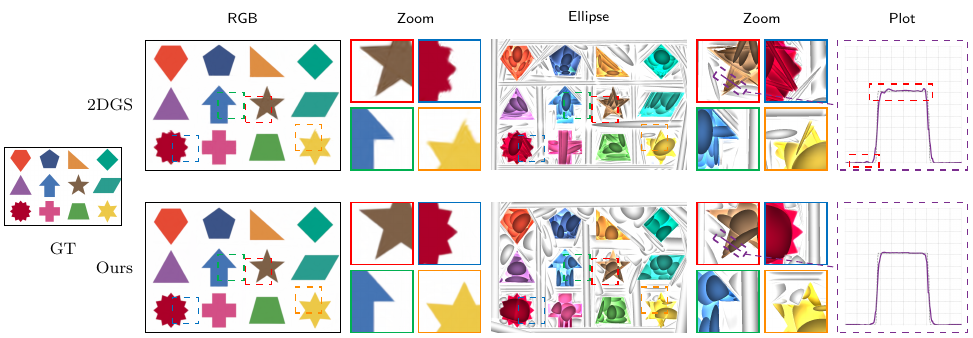}
   \caption{\textbf{Warped kernels suppress Gibbs.} Comparison between 2DGS (top) and Warp-2DGS (bottom) on a synthetic 2D scene. Dashed insets zoom into regions with sharp boundary features, where 2DGS suffers from Gibbs-like ringing artifacts caused by superimposing many Gaussian kernels to approximate discontinuities. Right (Plot): inverted gray value along the purple scanline. Our warped kernels can adopt a step-like edge region that suppresses such oscillatory overshoots.}
   \label{fig:teaser}
\end{figure}
\begin{abstract}
2D Gaussian Splatting (2DGS) builds on 3DGS with planar Gaussian primitives that improve surface reconstruction and geometric fidelity. However, the smooth falloff of Gaussian kernels limits the reconstruction of high-frequency shape signals, over-smoothing edges and fine geometric detail. Prior works replace the kernel function, yet our systematic evaluation on signal fitting shows that all existing kernels suffer from the Gibbs phenomenon, producing oscillatory artifacts near sharp discontinuities. We therefore make the kernel deformable instead of replacing it. Warp-2DGS augments each Gaussian surfel with a learnable, monotonic coordinate remapping in its local tangent plane. With one free control point per axis (four parameters per primitive), the remapping yields asymmetric radial profiles that capture sharp edges and irregular structures symmetric kernels cannot represent. It preserves the differentiable rasterization pipeline of 2DGS with moderate training overhead and is optimized end-to-end through the rendering loss alone. Experiments on MipNeRF\,360 and Tanks\,\&\,Temples show that our method suppresses Gibbs artifacts at sharp boundaries and achieves the best PSNR among 2DGS-based approaches, with consistent gains over 2DGS in PSNR, SSIM, and LPIPS. The advantage holds across all primitive budgets, at comparable or fewer Gaussians. Boundary PSNR improves over 2DGS at every dilation radius, and our method retains the mesh quality of 2DGS. Our project page is available at \href{https://matt-v50.github.io/Warp-2DGS/}{https://matt-v50.github.io/Warp-2DGS/}.
\end{abstract}
\section{Introduction}
\label{sec:intro}

Neural radiance fields have significantly advanced photorealistic novel view synthesis by encoding scenes as implicit representations. While 3D Gaussian Splatting (3DGS) introduced an explicit alternative using anisotropic 3D Gaussian primitives for real-time rendering, its volumetric nature often leads to noisy geometry and multi-surface artifacts. 2D Gaussian Splatting (2DGS) addresses these limitations by collapsing the primitives to planar, surfel-like 2D Gaussians, yielding more accurate surface reconstruction while preserving real-time performance. Its improved geometric fidelity and inherent surface alignment have driven growing adoption across applications that demand both rendering quality and reliable geometry~\cite{zhong2025Globally,yang2025Introducing,wang2024Spacetime,pak2025G2SICP,gao2025Gaussian,dai2024Highqualitya,yan20252DGSAvatar}.

Despite these advantages, the standard Gaussian kernel inherently lacks the ability to represent high-frequency shape signals. This limitation leads to over-smoothed reconstructions, particularly along sharp edges and fine geometric details. As illustrated in~\fref{fig:teaser}, existing methods exhibit visible artifacts in these regions, where the smooth Gaussian falloff fails to capture abrupt signal transitions.

Recent efforts have attempted to address this issue by replacing the Gaussian kernel with alternative primitives ~\cite{hamdi2024GES, liu2025Deformable, liu2025universal, zhu20253D, arunan2025DARBSplatting, chen2024Gaussians, li20253DHGS, kasymov2024NegGS, huang2025Deformable}. However, through systematic evaluation of a variety of existing kernel functions in signal fitting tasks, we observe that the problem remains challenging to fully resolve. In particular, all tested kernels give rise to the Gibbs~\cite{gottlieb1997gibbs} phenomenon when approximating sharp boundaries, producing characteristic oscillatory artifacts near discontinuities. This finding suggests that a fundamentally different kernel design is required — one that can faithfully represent sharp edge signals while remaining smooth enough to be differentiable and trainable. Furthermore, since shape signals in real scenes are spatially non-uniform, a fully symmetric kernel is inherently limited in its expressiveness.

To address these challenges, we propose a warped Gaussian kernel that provides asymmetric and adaptive shape representation while preserving differentiability for end-to-end optimization. The deformation is realized by composing the Gaussian with a learnable radial warping function defined by a set of free control points on the local tangent plane introduced by 2DGS, allowing each primitive to reshape its radial profile into arbitrary monotonically decreasing forms beyond the fixed exponential falloff. Operating directly on the 2D tangent plane avoids deforming a 3D volumetric kernel and projecting it onto the image plane, which would otherwise require costly numerical integration or approximation. Our contributions are summarized as follows:

\begin{itemize}[leftmargin=*]
    \item We introduce a warped Gaussian kernel that combines asymmetric learning capacity with differentiability, enabling sharp boundary representation that symmetric smooth kernels inherently cannot achieve.
    \item We demonstrate that our approach effectively mitigates the Gibbs phenomenon at sharp edges, a fundamental limitation inherent in existing kernel alternatives.
    \item Our method achieves the best PSNR among 2DGS-based approaches on standard benchmarks.
\end{itemize}
\section{Related Work}
\label{sec:related}

The standard Gaussian kernel enables efficient differentiable rasterization but its symmetric radial falloff struggles to represent sharp geometric features such as edges, occlusion boundaries, and thin structures. Real-world scenes often contain spatially asymmetric signals that a fixed symmetric kernel cannot faithfully capture with a single primitive. A growing body of work has sought to address this limitation by modifying the kernel function or replacing the primitive altogether ~\cite{hamdi2024GES, liu2025Deformable, liu2025universal, zhu20253D, arunan2025DARBSplatting, chen2024Gaussians, li20253DHGS, kasymov2024NegGS, huang2025Deformable}.

\textbf{Kernel function}. GES~\cite{hamdi2024GES} makes the exponent of the Gaussian function learnable, allowing each primitive to control its falloff sharpness. BetaGS~\cite{liu2025Deformable} replaces the Gaussian with a Beta distribution kernel that offers bounded support, adaptive frequency control, and asymmetric shape modeling, and its successor, Universal Beta Splatting~\cite{liu2025universal}, extends this to N-dimensional anisotropic settings. StudentGS~\cite{zhu20253D} adopts the Student-t distribution for heavier tails. DARB Splatting~\cite{arunan2025DARBSplatting} systematically compares a family of decaying anisotropic radial basis functions, including half-cosine, raised-cosine, and multiquadric kernels. 3DLS~\cite{chen2024Gaussians} replaces the exponential with a bounded linear kernel to sharpen primitive boundaries and reduce floaters. 3D-Half GS~\cite{li20253DHGS} splits each Gaussian into two halves with independent opacities, enabling one-sided sharp edges. NegGS~\cite{kasymov2024NegGS} introduces negative-amplitude Gaussians to counteract the long-tail effect through destructive interference. DRK~\cite{huang2025Deformable} enriches 2D Gaussian surfels with learnable radial basis functions defined by multiple free control points to represent arbitrary closed shapes. DRK changes the footprint angularly, whereas our warp changes the falloff along each tangent axis. The two act in complementary directions and could, in principle, be combined. However, despite their varied radial falloffs and asymmetric shapes, these kernels still exhibit the Gibbs phenomenon when fitting sharp edges or high-frequency content (see~\sref{ssec:kernel}).

\textbf{Gaussian primitive} with alternative geometric representations. 3DCS~\cite{held20253D} uses a smooth convex polyhedron to better resolve planar surfaces and edges. Triangle Splatting~\cite{held2026Triangle} optimizes explicit triangles, bridging the gap to traditional mesh-based rendering pipelines. BG-Triangle~\cite{wu2025BGTriangle} combines Bézier triangle surfaces with pixel-aligned Gaussian sub-primitives for vectorized scene representation with sharp boundary preservation. QGS~\cite{zhang2025Quadratic} extends the planar surfel to a quadratic surface model, enabling each primitive to capture local curvature for improved geometric fitting. These approaches change the geometry of the primitive itself and are thus complementary to kernel-level modifications. They do not address the limited expressiveness of the falloff profile within a planar primitive, which is why our main comparison focuses on methods that retain a planar primitive.

Our approach is orthogonal to these methods. Rather than modifying the kernel function or replacing the primitive, we introduce a learnable coordinate remapping in the local $u-v$ tangent plane of each 2D Gaussian surfel. The Gaussian kernel itself remains unchanged; instead, the input coordinates are passed through a monotonic function before evaluation, enabling asymmetric radial falloff with minimal additional parameters. Since our method operates on the input coordinates through a learnable radial mapping rather than replacing the kernel function itself, it offers a more general formulation that can recover similar effective profiles under suitable parameterizations.
\section{Method}
\label{sec:method}

\subsection{Preliminaries: 2D Gaussian Splatting}
\label{sec:2dgs}
Gaussian Splatting methods represent scenes using collections of Gaussian primitives, typically initialized from point clouds obtained via Structure-from-Motion (SfM)~\cite{schoenberger2016sfm} or LiDAR. While 3DGS~\cite{kerbl3Dgaussians} uses anisotropic 3D ellipsoids, its volumetric nature often leads to view-inconsistent geometry. 2DGS~\cite{huang20242D} addresses this by reformulating each primitive as a 2D oriented planar disk embedded in 3D space, enabling more accurate surface reconstruction. Each 2D Gaussian is parameterized by a set of learnable attributes: center $\boldsymbol{\mu} \in \mathbb{R}^3$, two principal tangent vectors $\mathbf{t}_u, \mathbf{t}_v \in \mathbb{R}^3$ spanning the local surface plane, corresponding scaling parameters $s_u, s_v \in \mathbb{R}$, opacity $o \in \mathbb{R}$, and color represented via spherical harmonics. A point in world space is parameterized in the local tangent frame as:
\begin{equation}
\mathbf{x}(u, v) = \boldsymbol{\mu} + s_u \mathbf{t}_u \, u + s_v \mathbf{t}_v \, v,
\end{equation}
and the 2D Gaussian is evaluated as:
\begin{equation}
\mathcal{G}(u,v) = e^{-\frac{1}{2}(u^2 + v^2)},
\end{equation}
where $(u,v)$ are the local tangent-plane coordinates of the ray--surfel intersection. The pixel color $\mathbf{C}(\mathbf{p})$ is computed by alpha-compositing the ordered set of overlapping Gaussians:
\begin{equation}
\mathbf{C}(\mathbf{p}) = \sum_{i=1} \mathbf{c}_i \, \alpha_i \, \hat{\mathcal{G}}_i(u_i,v_i) \prod_{j=1}^{i-1} \left(1 - \alpha_j \, \hat{\mathcal{G}}_j(u_j,v_j)\right),
\end{equation}
where $\mathbf{c}_i$ is the view-dependent color, $\alpha_i$ is the learned per-Gaussian opacity, and $\hat{\mathcal{G}}$ is a low-pass filtered kernel defined as $\hat{\mathcal{G}} = \max\!\left\{\mathcal{G}(u,v),\, \mathcal{G}_{2d}(\mathbf{p})\right\}$, where $\mathcal{G}_{2d}$ is a screen-space Gaussian filter that prevents aliasing at distant or oblique views. All attributes are optimized end-to-end through differentiable rasterization with photometric supervision from multi-view images. 

\subsection{Kernel function}
\label{ssec:kernel}

The GS optimization processing could be seen as a signal fitting process. Unlike the transformation of frequency domain, GS does not have a complete basis functions such as the Fourier series and the wavelet transform. The primitive densify processing, split and clone, has to use smaller and narrower ellipsoids to compensate for the high frequency signal, which is the most important component of the spatial domain. However, the accumulation of those different-sized and scaled ellipsoids leads to the Gibbs phenomenon. The proposed warped Gaussian kernel reshapes the kernel profile along each principal axis via a learnable warp function, enabling sharp boundary representation without Gibbs artifacts.

\begin{figure}[htbp]
  \centering
  \includegraphics[width=1\textwidth]{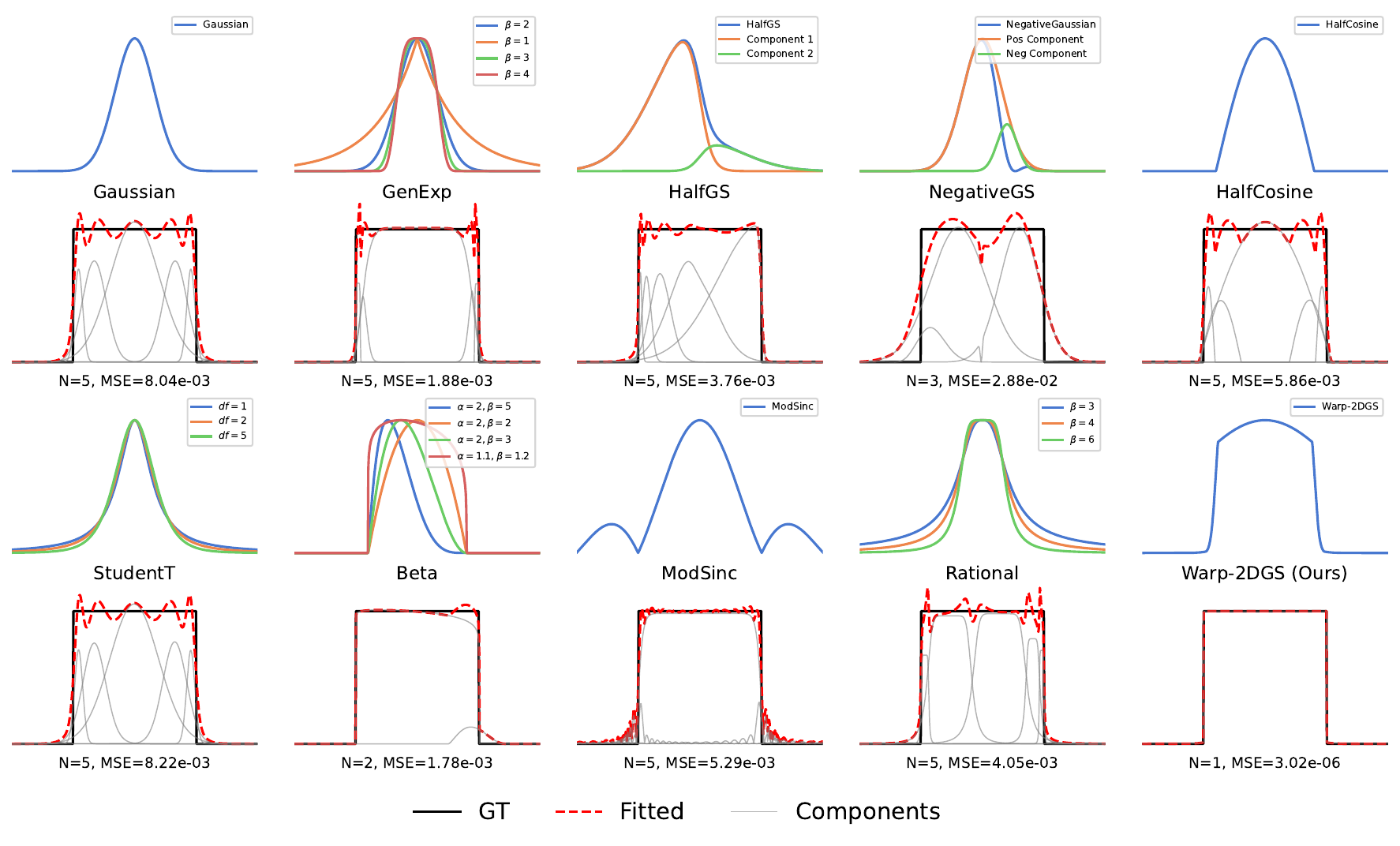}
   \caption{\textbf{Kernel function comparison and fitting results}. Odd rows show the primitives using different kernel functions, and even rows show the corresponding best-fitting results under 5 components. The red dashed line indicates the fitted curve. For our kernel, this 1D illustration uses a center-symmetric variant with two mirrored control points, whereas all reconstruction experiments use the single-knot kernel of~\sref{ssec:deformation}.}
   \label{fig:gibbs}
\end{figure}

As shown in~\fref{fig:gibbs}, we fit a 1D square wave defined as $f(x) = \mathbf{1}_{[0.25,\, 0.75]}(x)$ using various kernel functions under a simplified 1D Gaussian reconstruction setting, which can be interpreted as observing the signal through a fixed-position orthographic camera from the top center, where contributions from up to 5 components are directly summed without alpha blending. For specific square wave fitting algorithms and all results, please refer to~\aref{app:algorithm}. For each kernel type, we greedily search for the best combination rather than relying on clone-and-split. The results show that, except for the Beta function, all kernels tend to use the maximum number of components, and the Gibbs phenomenon persists across all tested kernels.

To prevent this phenomenon from happening, we choose a deformable mapping method to stretch the original Gaussian function to fit the square wave, where the sharp edges of it frequently appear in the scenes that need to be reconstructed.

\subsection{Kernel Warping}

The 1D deformed Gaussian is defined as $G(D(x)) = e^{-\frac{1}{2}D(x)^2}$. We extend this to 2D with axis-separated deformations $D_u(u)$ and $D_v(v)$ to avoid cross-axis distortion:
\begin{equation}
G(D(u, v)) = e^{-\frac{1}{2}(D_u(u)^2 + D_v(v)^2)}.
\end{equation}

The design of this deformable mapping function should satisfy the following condition:

\begin{enumerate*}[label=(\arabic*)]
  \item The deformed function should remain smooth to maintain learnability like the original Gaussian,
  \item the deformed function should not distort the Gaussian function too much, which would cause the function to introduce high-frequency fluctuations prematurely, and
  \item the function should be able to recover the original Gaussian function during the training process. In other words, $D(t|\theta)\equiv t$ for some parameter setting.
\end{enumerate*}

\begin{figure}[htbp]
  \centering
  \includegraphics[width=0.8\textwidth]{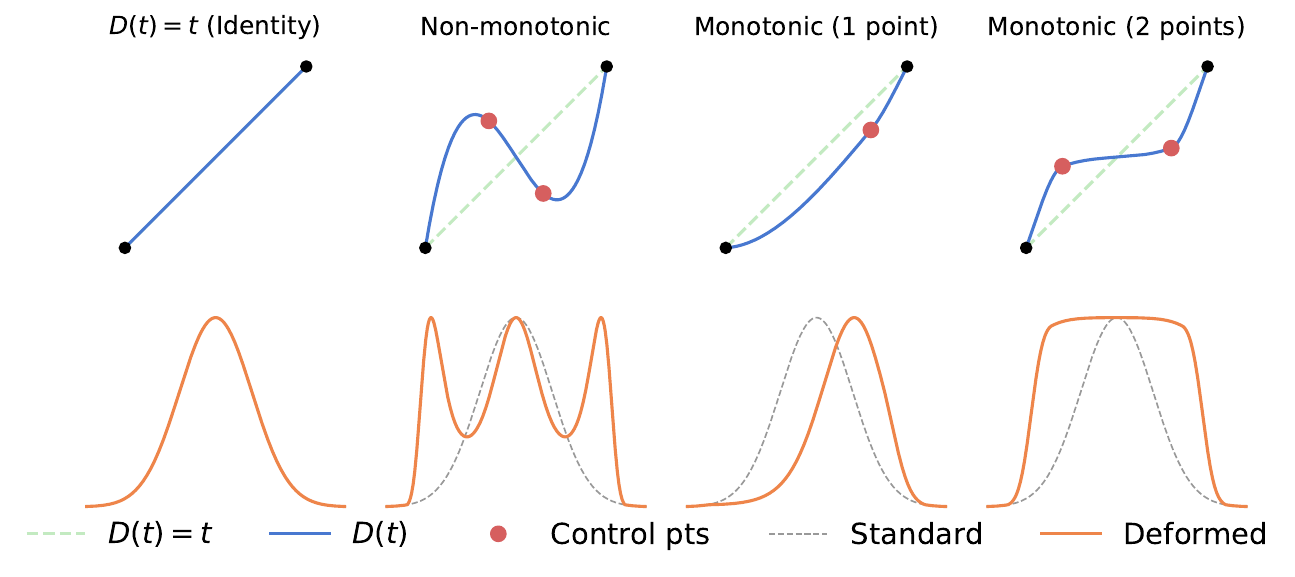}
   \caption{\textbf{Deformation functions and their corresponding deformed Gaussian kernels.} Top row: deformation function $D(t)$ with $t = x/r$. Bottom row: the resulting deformed Gaussian $G(r \cdot D(x/r))$, where $r$ is a rescaling factor such that one unit in $t$ corresponds to $r$ standard deviations in the original domain.}   \label{fig:deformation}
\end{figure}

Under these conditions, the simplest deformation function $D(t)$ should be monotonically increasing within the given bounds. As shown in~\fref{fig:deformation}, we compare the effects of different deformation functions on the Gaussian kernel. The rescaling factor $r$ normalizes the coordinate as $t = x/r$, so that one unit in $t$-space corresponds to $r$ standard deviations of the Gaussian, allowing the deformation function to operate on a standardized scale range. A non-monotonic deformation produces multiple peaks in the deformed kernel, which not only can be equivalently represented by a mixture of standard Gaussians — undermining the shape--scale decoupling — but also introduces Gibbs-like ringing artifacts that our method is specifically designed to suppress. The last two columns demonstrate that adding more control points enables fitting higher-frequency signals, where the control points guide the model to capture sharp transitions while reducing overshooting near edges.

\subsection{Deformation Function}
\label{ssec:deformation}
The way to construct the deformation function is to use a
piecewise linear (PWL) monotonic function. The endpoints are anchored at $(-1,-1)$ and $(1,1)$ to preserve the identity mapping at the boundary. A single learnable interior knot $(c_1,d_1=D(c_1))$ then fully specifies the piecewise-linear warp through three control points:
\begin{equation}
  \{(c_i, d_i)\}_{i=0}^{2}
  = \bigl\{(-1,-1),\;(c_1,d_1),\;(1,1)\bigr\}.
  \label{eq:control_points}
\end{equation}

with $-1 < c_1 < 1$ and $-1 < d_1 < 1$. The deformation function is
\begin{equation}
D(t|\mathbf{c},\mathbf{d}) =
\begin{cases}
t, & |t| > 1, \\[4pt]
d_i + \dfrac{d_{i+1} - d_i}{c_{i+1} - c_i}\,(t - c_i),
& t \in [c_i,\, c_{i+1}], \quad i = 0, 1.
\end{cases}
\label{eq:piecewise_monotonic}
\end{equation}

\paragraph{Forward pass.}
For $t \in [c_i, c_{i+1}]$, define the interval length $h_i$, the vertical displacement $\Delta d_i$, and the local coordinate $o$:
\[
h_i = c_{i+1} - c_i, \qquad \Delta d_i = d_{i+1} - d_i, \qquad o = \frac{t - c_i}{h_i}.
\]
Then the PWL interpolation can be written as
\begin{equation}
D(t) = (1-o)d_i + o d_{i+1}.
\end{equation}

\paragraph{Local backward pass.}
The gradients with respect to $t$ and the interior control point $(c_1, d_1)$ are:
\begin{equation}
\frac{\partial D}{\partial t} = \frac{\Delta d_i}{h_i}, \quad
\frac{\partial D}{\partial c_1}
=
\begin{cases}
-o\, \frac{\Delta d_0}{h_0}, & t \in [c_0, c_1], \\[6pt]
-(1-o)\,\frac{\Delta d_1}{h_1}, & t \in [c_1, c_2],
\end{cases}
\quad
\frac{\partial D}{\partial d_1}
=
\begin{cases}
o, & t \in [c_0, c_1], \\[6pt]
1-o, & t \in [c_1, c_2].
\end{cases}
\label{eq:pwl_backward}
\end{equation}

These partial derivatives are independent of how the interior control point is parameterized. Different parameterizations only change the chain rule from the learnable parameters to $(c_1, d_1)$.

\subsubsection{Direct Parameterization}

A direct parameterization learns the horizontal and vertical coordinates of the interior control point independently. Let $a,b \in \mathbb{R}$ be two learnable scalars. We map them to $(-1,1)$ by
\begin{equation}
c_1 = 2\sigma(a) - 1,
\qquad
d_1 = 2\sigma(b) - 1,
\end{equation}
where $\sigma(\cdot)$ denotes the sigmoid function. This automatically
ensures $-1 < c_1 < 1$ and $-1 < d_1 < 1$ without sorting or clamping.

The chain rule gives
\begin{equation}
  \frac{\partial G}{\partial a} = \frac{\partial G}{\partial \tilde t}\cdot\frac{\partial \tilde t}{\partial c_1}\cdot\frac{\partial c_1}{\partial a},
  \qquad
  \frac{\partial G}{\partial b} = \frac{\partial G}{\partial \tilde t}\cdot\frac{\partial \tilde t}{\partial d_1}\cdot\frac{\partial d_1}{\partial b}
  \label{eq:chain_direct}
\end{equation}
where $\tilde t = r\,D(t/r)$ is the warped coordinate, so that
\begin{equation}
  \frac{\partial \tilde t}{\partial a} = r\,\frac{\partial D(t/r)}{\partial c_1}\,\frac{\partial c_1}{\partial a},
  \qquad
  \frac{\partial \tilde t}{\partial b} = r\,\frac{\partial D(t/r)}{\partial d_1}\,\frac{\partial d_1}{\partial b}.
  \label{eq:grad_direct}
\end{equation}
\subsubsection{Slope-Ratio Parameterization}


\definecolor{ptA}{HTML}{FF0000}  
\definecolor{ptB}{HTML}{0070C0}  
\definecolor{ptC}{HTML}{00B050}  
\definecolor{ptD}{HTML}{FF8C00}  

\newcommand{\imgwidth}{0.108\linewidth}
\newcommand{\imgwidthsmall}{0.09\linewidth}
\newcommand{\methodA}{2DGS_dev_pwl3_x2_3sigma_2grad_fixed_again_sqrt2_fixed_inter_lr_0.0025}
\newcommand{\methodB}{2DGS_dev_pchip3_x2_3sigma_grad_fixed_sqrt2_fixed_inter_lr_0.0025}
\newcommand{\methodC}{2DGS_dev_pwl3_pq_x2_3sigma_sqrt2_fixed_inter_lr_0.0025}
\newcommand{\methodD}{2DGS_dev_pchip3_pq_x2_3sigma_sqrt2_fixed_inter_lr_0.0025}
\newcommand{\methodAlabel}{PWL}
\newcommand{\methodBlabel}{Pchip}
\newcommand{\methodClabel}{PWL \textbf{sr}}
\newcommand{\methodDlabel}{Pchip \textbf{sr}}

\newcommand{\scenerow}[2]{%
  \scene{\detokenize\expandafter{#1}/#2/bonsai_c}  &
  \scene{\detokenize\expandafter{#1}/#2/counter_c} &
  \scene{\detokenize\expandafter{#1}/#2/kitchen_c} &
  \scene{\detokenize\expandafter{#1}/#2/room_c}    &
  \scene{\detokenize\expandafter{#1}/#2/garden_c}  &
  \scene{\detokenize\expandafter{#1}/#2/stump_c}   &
  \scene{\detokenize\expandafter{#1}/#2/bicycle_c}
}

\newcommand{\scenerowsix}[2]{%
  \scene{\detokenize\expandafter{#1}/#2/bonsai_c}  &
  \scene{\detokenize\expandafter{#1}/#2/counter_c} &
  \scene{\detokenize\expandafter{#1}/#2/kitchen_c} &
  \scene{\detokenize\expandafter{#1}/#2/room_c}    &
  \scene{\detokenize\expandafter{#1}/#2/garden_c}  &
  \scene{\detokenize\expandafter{#1}/#2/stump_c}
}

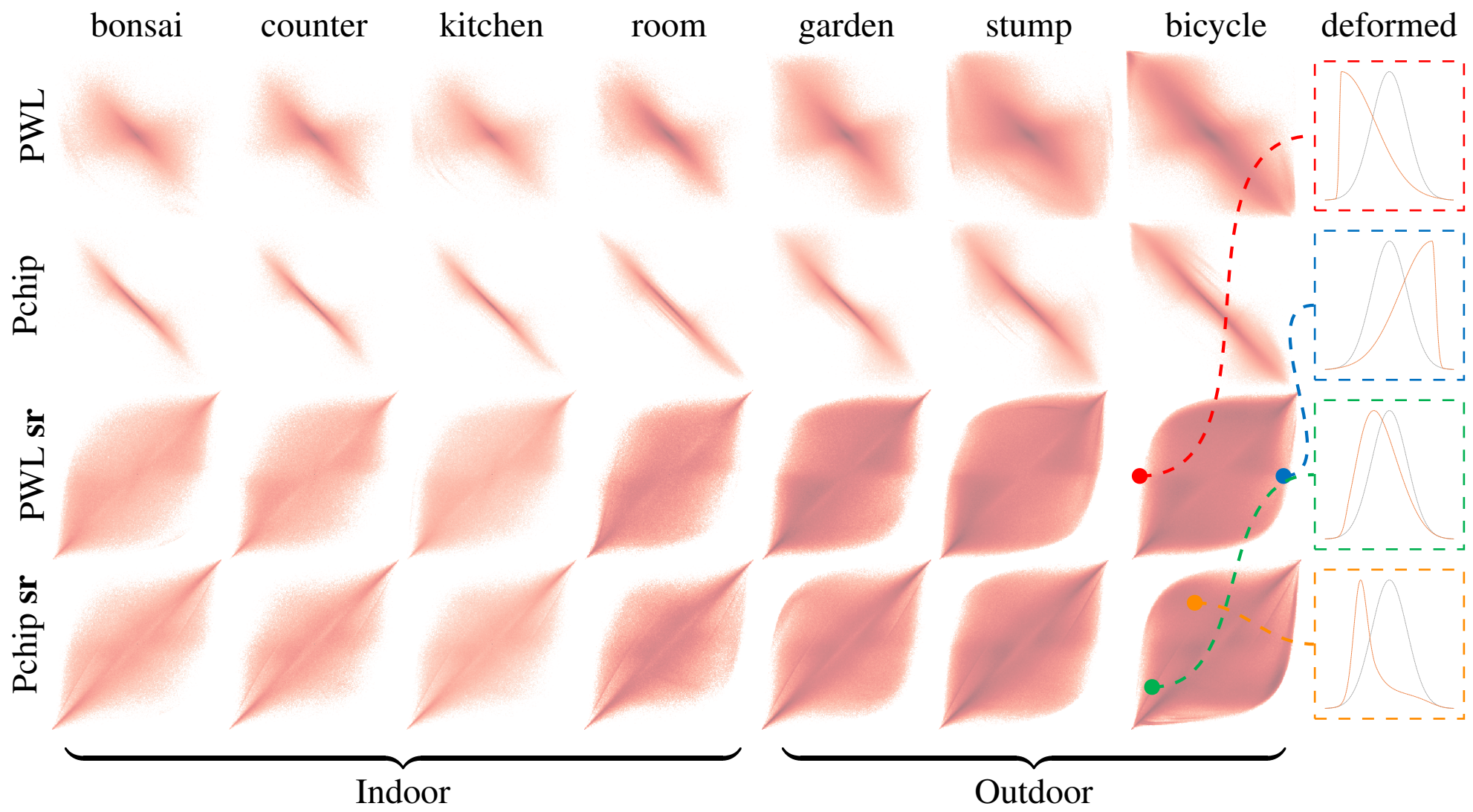
\begin{figure}[t]
  \centering
  \setlength{\tabcolsep}{1pt}
  \renewcommand{\arraystretch}{0.5}

  \newcommand{\scene}[1]{%
    \begin{tikzpicture}[baseline={(current bounding box.center)}]
      \node[inner sep=0] {%
        \includegraphics[width=\imgwidth]{main/fig/methods/point_distribution/#1}};
    \end{tikzpicture}%
  }

  \newcommand{\scenenamed}[2]{
    \begin{tikzpicture}[remember picture, baseline={(current bounding box.center)}]
      \node[inner sep=0] (#2) {%
        \includegraphics[width=\imgwidth]{main/fig/methods/point_distribution/#1}};
    \end{tikzpicture}%
  }

    \newcommand{\explimg}[3]{
        \begin{tikzpicture}[remember picture, baseline={(current bounding box.center)}]
          \node[inner sep=1pt, draw=#2, line width=0.5pt, dashed] (#1) {%
            \includegraphics[width=\imgwidthsmall, height=\imgwidthsmall]{main/fig/methods/point_distribution/deform/#3}};
        \end{tikzpicture}%
    }

  \begin{tabular}{r*{7}{c}c}
    & \small bonsai  & \small counter & \small kitchen & \small room
    & \small garden  & \small stump   & \small bicycle & \small deformed\\[2pt]
    %
    $\vcenter{\hbox{\rotatebox{90}{\small \methodAlabel{}}}}$ &
    \scenerow{\methodA}{x8} &
    \explimg{explA}{ptA}{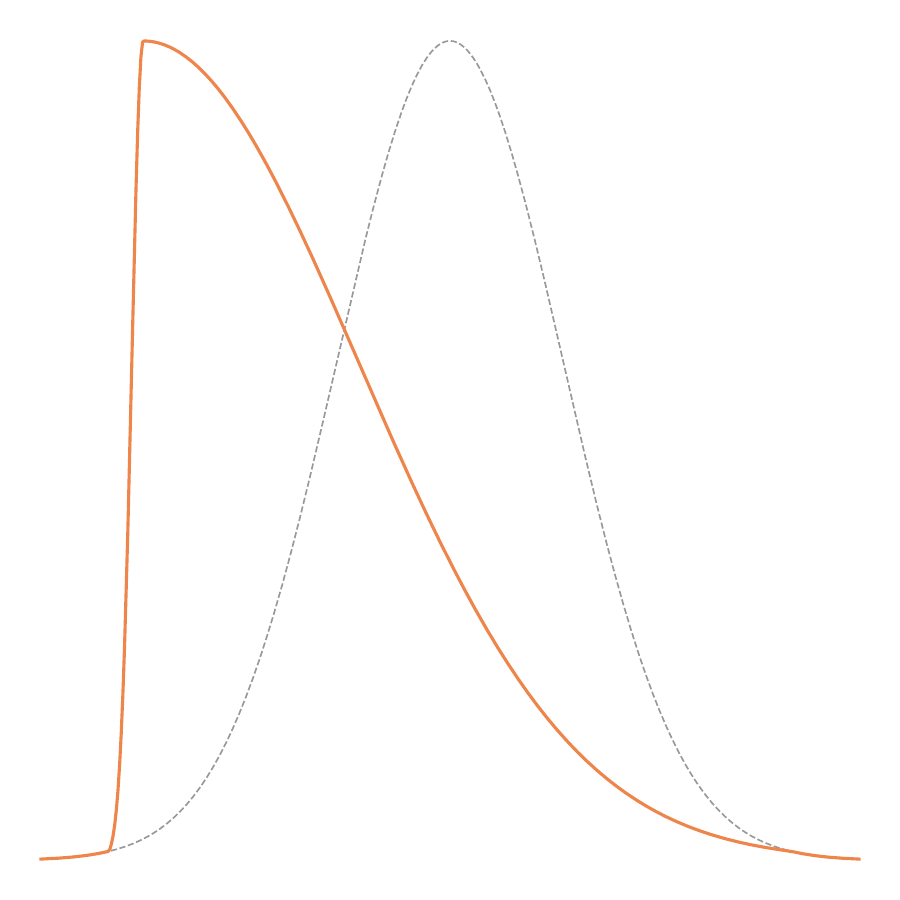} \\[2pt]
    %
    $\vcenter{\hbox{\rotatebox{90}{\small \methodBlabel{}}}}$ &
    \scenerow{\methodB}{x8} &
    \explimg{explB}{ptB}{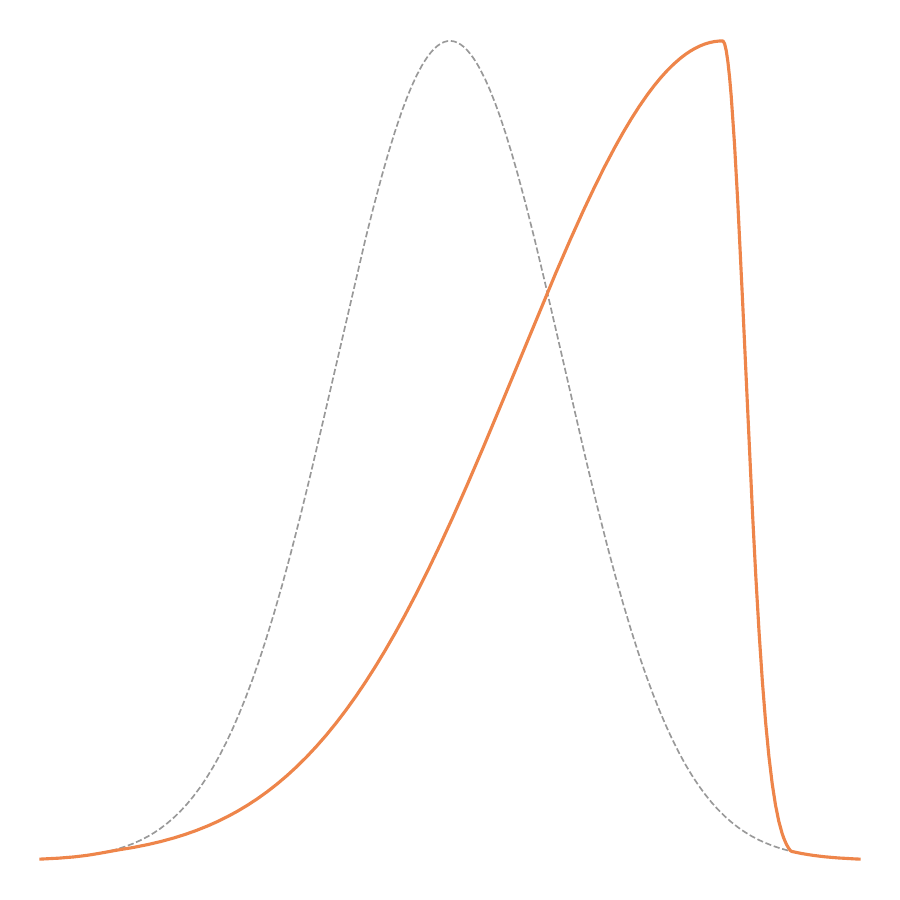} \\[2pt]
    %
    $\vcenter{\hbox{\rotatebox{90}{\small \methodClabel{}}}}$ &
    \scenerowsix{\methodC}{x8} &
    \scenenamed{\detokenize\expandafter{\methodC}/x8/bicycle_c}{pwlsrbic} &
    \explimg{explC}{ptC}{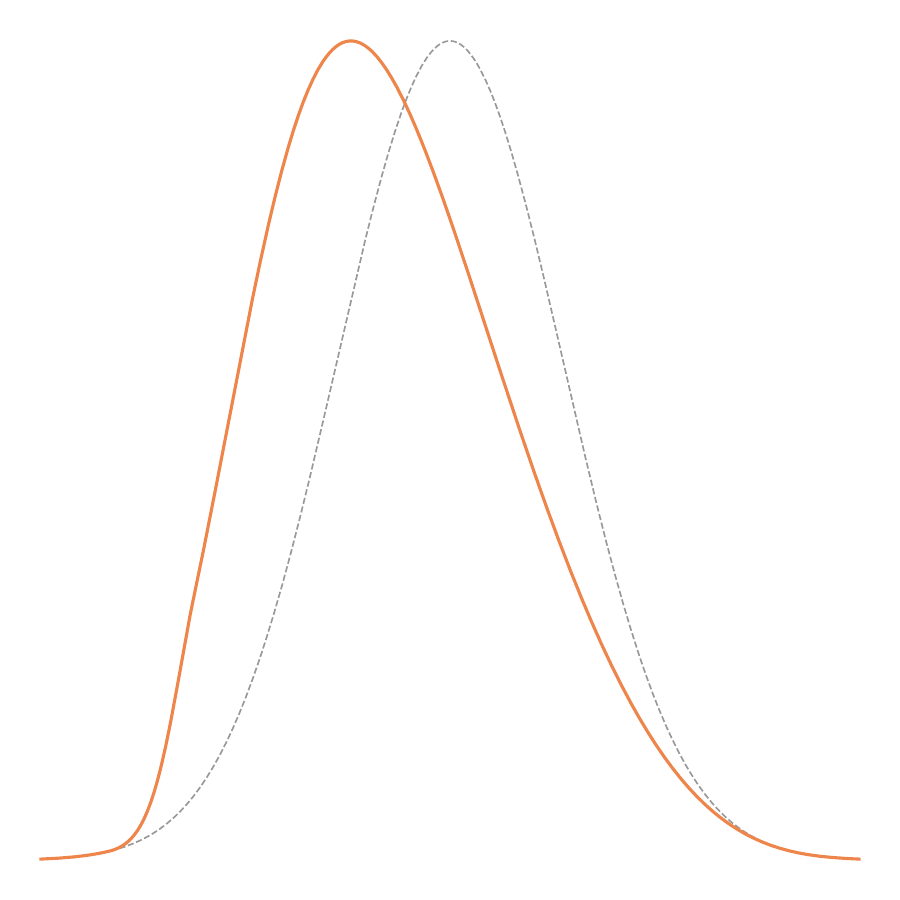} \\[2pt]
    %
    $\vcenter{\hbox{\rotatebox{90}{\small \methodDlabel{}}}}$ &
    \scenerowsix{\methodD}{x8} &
    \scenenamed{\detokenize\expandafter{\methodD}/x8/bicycle_c}{pchipsrbic} &
    \explimg{explD}{ptD}{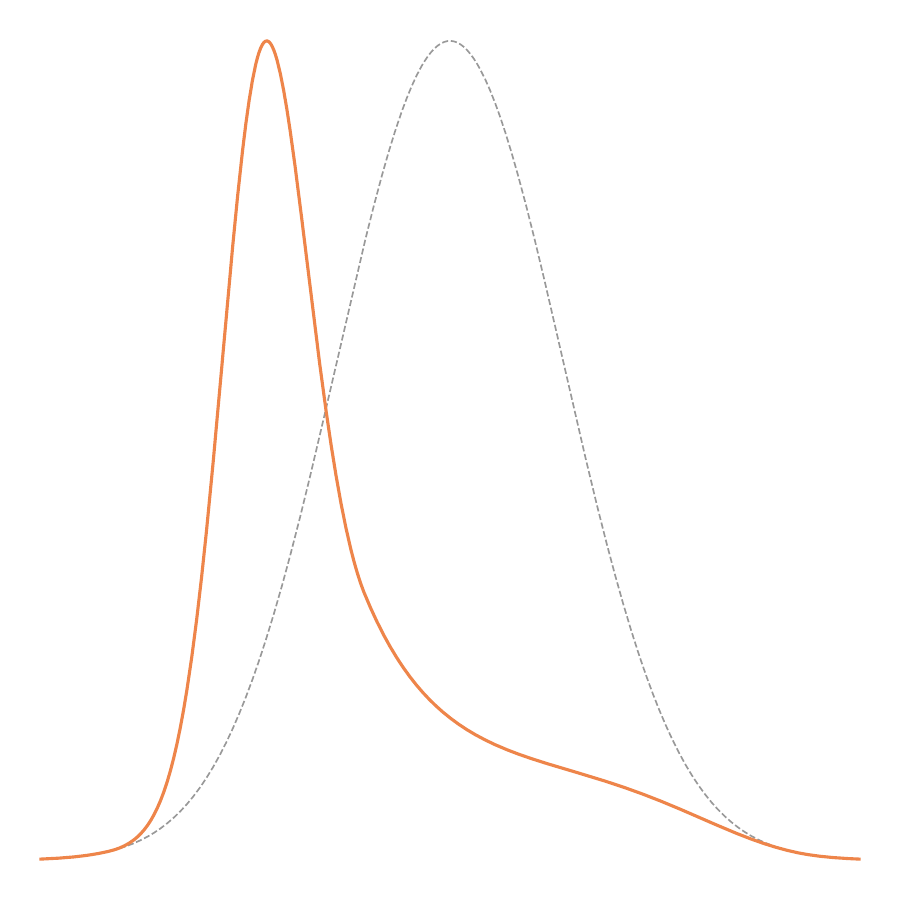} \\[2pt]
    &
    \multicolumn{4}{c}{$\underbrace{\hphantom{\rule{0.43\linewidth}{0pt}}}_{\text{\small Indoor}}$} &
    \multicolumn{3}{c}{$\underbrace{\hphantom{\rule{0.32\linewidth}{0pt}}}_{\text{\small Outdoor}}$} &
    \\[-4pt]
  \end{tabular}

  \begin{tikzpicture}[remember picture, overlay]
    \fill[ptA] ([xshift=0.8mm, yshift=7.5mm]pwlsrbic.south west)
        circle (2pt) coordinate (mA);
    \fill[ptB] ([xshift=13.6mm, yshift=7.5mm]pwlsrbic.south west)
        circle (2pt) coordinate (mB);
    \fill[ptC] ([xshift=1.9mm, yshift=3.8mm]pchipsrbic.south west)
        circle (2pt) coordinate (mC);
    \fill[ptD] ([xshift=5.7mm, yshift=11.3mm]pchipsrbic.south west)
        circle (2pt) coordinate (mD);

    \draw[ptA, dashed, thick] (mA) to[out=0, in=180] (explA.west);
    \draw[ptB, dashed, thick] (mB) to[out=0, in=180] (explB.west);
    \draw[ptC, dashed, thick] (mC) to[out=0, in=180] (explC.west);
    \draw[ptD, dashed, thick] (mD) to[out=0, in=180] (explD.west);
  \end{tikzpicture}

  \caption{\textbf{Control point distributions across MipNeRF 360 scenes.} 2D histogram of $(c,d)$  coordinates along the $u$ direction across all scenes in the MipNeRF~360 dataset ($\times 8$ downsampling). The \textbf{sr} suffix denotes the slope-ratio parameterization. Colored markers highlight representative $(c,d)$ positions; dashed boxes on the right show the corresponding deformed Gaussian kernels (orange) against the standard Gaussian (gray).}
  \label{fig:point_distribution}
\end{figure}

Although the direct control-point parameterization is straightforward, it can be unstable during optimization because the horizontal and vertical coordinates of the interior control points are strongly coupled through direct parameterization. In particular, for an increasing monotone deformation, the local gradients with respect to $c_1$ and $d_1$ tend to have opposite signs. This can be seen directly from the local gradient formulas above \eqref{eq:pwl_backward}. The two gradients $\frac{\partial D}{\partial c_1}, \frac{\partial D}{\partial d_1}$ are linearly coupled by the positive local slope $\Delta d_i/h_i$ and therefore have opposite signs; near the origin, they are nearly negatives of each other, causing the control point to move predominantly along the anti-diagonal direction during gradient descent. \fref{fig:point_distribution} visualizes the distribution of learned control points in their parameter space. A broader spread indicates greater adaptability, while concentration along the anti-diagonal implies that the control points are strongly correlated, effectively restricting deformations to a low-dimensional subspace. Outdoor scenes consistently exhibit a wider spread than indoor ones, reflecting higher demand for complex kernel shapes. The sampled examples at the extremes further reveal asymmetric deformations associated with higher-frequency representations.

Moreover, the sensitivity is amplified when one of the PWL intervals becomes short. The segment length $h_i$ appears in the denominator of the segment slope, and the gradient with respect to the knot location contains an even stronger dependence on the interval length, $\frac{\partial D}{\partial c_1}\propto\frac{\Delta d_i}{h_i^2}$. To decouple these effects, we instead parameterize the PWL deformation by the knot location $c$ and the slope ratio $R=\frac{d_0}{h_0}/\frac{d_1}{h_1}$ between the two segments. Let $a,b \in \mathbb{R}$ be two learnable scalars and define
\begin{equation}
c_1 = 2\sigma(a)-1,
\qquad
q = Q_{\max}\cdot(2\sigma(b) - 1),
\qquad
R = \exp(q)
\end{equation}

where $Q_{max}$ is set to control the maximum slope ratio, preventing excessively large slopes from causing numerical instability. An alternative would be to clamp the slope directly, but this introduces asymmetric gradients for $a$ and $b$, effectively requiring different learning rates for the two parameters. To satisfy the endpoint constraints $D(-1)=-1$ and $D(1)=1$, the slopes must satisfy
\begin{equation}
k_0 h_0 + k_1 h_1 = 2,
\qquad
k_0 = R k_1.
\end{equation}
Solving these constraints gives the induced interior control point
\begin{equation}
d_1 = -1 + \frac{2R h_0}{h_1 + R h_0}.
\end{equation}

The backward pass reuses the local PWL gradients derived above. We first compute gradients with respect to the induced control point $(c_1,d_1)$, and then apply the chain rule.

\subsection{Smooth Monotonic Deformation Function}

Smooth, continuously differentiable functions are generally preferred in deep learning for their well-behaved gradient flow. We therefore implement a smooth variant using Pchip (Piecewise Cubic Hermite Interpolating Polynomial)~\cite{fritsch1980monotone} as the interpolator. Pchip is parameterized by the same set of control points as PWL, but replaces the linear segments with cubic polynomials that provide $C^1$ continuity and smoother gradient flow during backpropagation, while guaranteeing monotonicity by construction through the Fritsch--Carlson condition~\cite{fritsch1980monotone}.

\paragraph{Pchip forward pass.}
Using the interval lengths $h_i$ and slopes $k_i$, the derivatives at the two fixed endpoints are:
\begin{equation}
m_0 = \max\!\left(\frac{(2h_0 + h_1)\,k_0 - h_0\,k_1}{h_0 + h_1},\; 0\right), \qquad
m_2 = \max\!\left(\frac{(2h_1 + h_0)\,k_1 - h_1\,k_0}{h_1 + h_0},\; 0\right).
\end{equation}

The derivative at the interior point $c_1$ is the weighted harmonic mean of the adjacent slopes:
\begin{equation}
m_1 =
\begin{cases}
\dfrac{(w_1+w_2)k_0k_1}{w_1k_1+w_2k_0}, & \text{if } k_0 k_1 > 0, \\[8pt]
0, & \text{otherwise},
\end{cases}
\end{equation}
where $w_1 = 2h_1 + h_0$ and $w_2 = h_1 + 2h_0$ denote the corresponding weights. The output is:
\begin{equation}
D(t) = H_0(o)\,d_i + H_1(o)\,h_i\,m_i + H_2(o)\,d_{i+1} + H_3(o)\,h_i\,m_{i+1},\quad t \in [x_i, x_{i+1}]
\end{equation}
where the cubic Hermite basis functions are:
\begin{align}
H_0(o) &= 2o^3 - 3o^2 + 1, &\quad H_1(o) &= o^3 - 2o^2 + o, \nonumber \\
H_2(o) &= -2o^3 + 3o^2,     &\quad H_3(o) &= o^3 - o^2.
\end{align}
Outside the interpolation range, $D(t) = t$ for $|t| > 1$. The full derivation of the backward pass for Pchip is provided in~\aref{app:pchip_backward}.

All these deformation functions modify only the 3D projected distance in $\mathcal{G}(u,v)$ for opacity calculation; depth is always computed from undeformed coordinates. Since the low-pass filter selects $\hat{\mathcal{G}} = \max\!\left\{\mathcal{G}(u,v),\, \mathcal{G}_{2d}(\mathbf{p})\right\}$ and $\mathcal{G}_{2d}$ is evaluated in screen space, the deformation is naturally bypassed whenever the filter activates.

\subsection{Extent Computation}
\begin{figure}[htbp]
  \centering
  \includegraphics[width=1\textwidth, trim=0 70 0 250, clip]{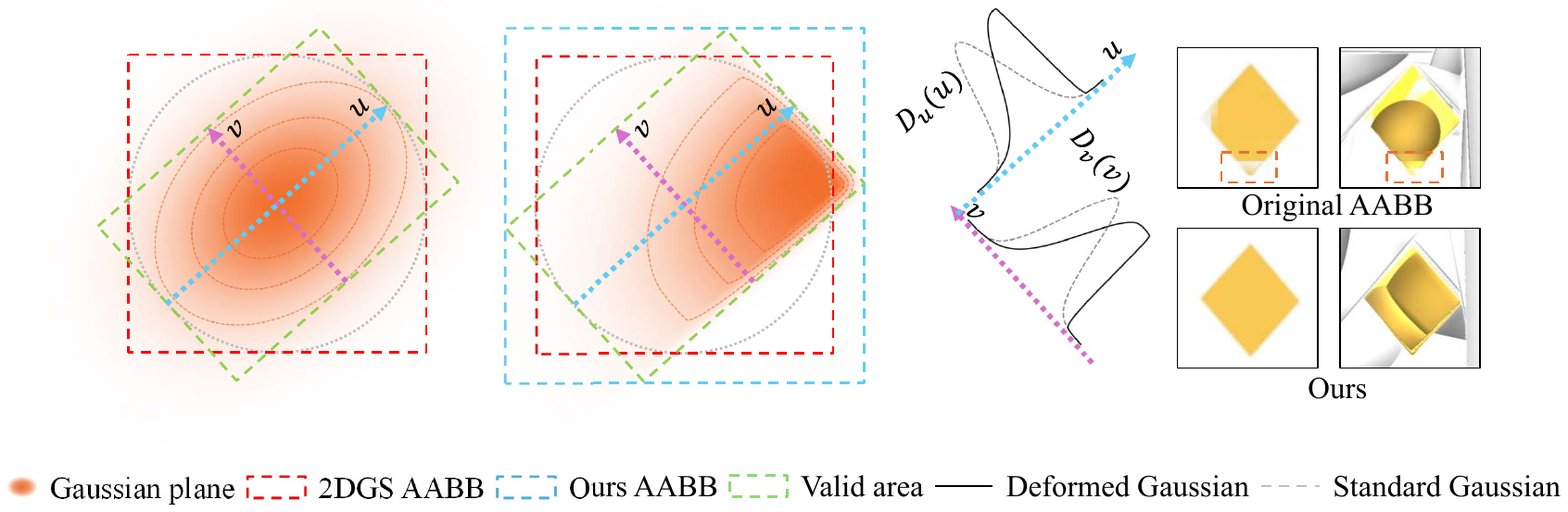}
   \caption{\textbf{AABB expansion for deformed kernels}. Left: The original 2DGS AABB (red) may clip the deformed Gaussian (orange); Middle: our expanded AABB (blue) fully covers the valid area (green). Right: deformation curves along the two principal axes, with zoomed-in comparisons of rendering with and without our AABB expansion.}
\label{fig:cutoff}
\end{figure}

In the rasterization stage~\cite{huang20242D}, the original 2DGS computes the Axis-Aligned Bounding Box (AABB) from the maximum axis of the projected ellipse. This relies on the assumption that values beyond the $3\sigma$ boundary are negligible. However, this assumption no longer holds when the Gaussian kernel is deformed. As illustrated in~\fref{fig:cutoff}, in the extreme case where $D_u(u) = D_v(v) = 3$, the Mahalanobis distance at the AABB corner remains $3$, leaving a non-negligible kernel value. Therefore, the AABB must be expanded to account for the deformation. Specifically, to prevent clipping artifacts, the cutoff value in this study should be multiplied by at least $\sqrt{2}$. 
\section{Experiments}
\label{sec:experiment}

\subsection{Experimental Setup}
All experiments are conducted on a single NVIDIA RTX 4090 GPU. We train for 30,000 iterations following the 2DGS training schedule with default hyperparameters. For our method, the deformation parameters use a rescale factor $r=3$ and a learning rate of $0.0025$, set to half of the scaling learning rate. $Q_{max}$ is set to 3 to prevent computational overflow, $\exp(Q_{max}) \simeq 20$. We initialize $a=b=0$ so that $\sigma(a)=\sigma(b)=0.5$. This gives $c_1=d_1=0$ and $R=1$, and the deformation function reduces to the identity under both parameterizations. Same as 2DGS, we use the same optimizer, Adam~\cite{kingma2014adam}, for other original parameters. To avoid affecting densification, we compute the radius using the original cutoff value $3$ for the densification strategy, but use the extended cutoff $3*\sqrt{2}$ for rasterization. For all baselines, we use their official implementations with default settings.

\subsection{Main Comparison}

We compare our method against several 2D-based Gaussian splatting methods: 2DGS~\cite{huang20242D}, TriangleGS~\cite{held2026Triangle}, and DRK-GS~\cite{huang2025Deformable} (evaluated under three densification settings: dense, middle, and sparse, as defined in their original work). We also include 3D-based methods 3DHGS~\cite{li20253DHGS} and BetaGS~\cite{liu2025Deformable} as reference. All methods are evaluated on MipNeRF 360~\cite{barron2022MipNeRF} ($\times 4$, $\times 8$) and Tanks\,\&\,Temples~\cite{knapitsch2017tanks} (COLMAP-processed by PixelGS~\cite{zhang2024PixelGS}), reporting PSNR, SSIM, and LPIPS.

\definecolor{best}{HTML}{FFCCCC}
\definecolor{second}{HTML}{FFE5CC}
\begin{table}[htbp]
\centering
\caption{\textbf{Quantitative comparison on the MipNeRF 360 and Tanks\,\&\,Temples datasets.} Outdoor and Indoor splits are averaged over their respective scenes. \colorbox{best}{best} and \colorbox{second}{second} among 2DGS methods. \textbf{sr}: slope parameterization method. $^\dagger$: 3DGS methods. * Results taken from the original paper. For DRK-GS, only dense is reported on TNT. 3DHGS omitted as full-scene results are unavailable.}
\label{tab:main_comparison}
\setlength{\tabcolsep}{2.8pt}
\resizebox{\columnwidth}{!}{%
\begin{tabular}{l|ccc|ccc|ccc|ccc|ccc}
\toprule
& \multicolumn{6}{c|}{MipNeRF360 $\times 4$} & \multicolumn{6}{c|}{MipNeRF360 $\times 8$} & \multicolumn{3}{c}{TNT $\times 1$} \\
& \multicolumn{3}{c|}{Outdoor} & \multicolumn{3}{c|}{Indoor} & \multicolumn{3}{c|}{Outdoor} & \multicolumn{3}{c|}{Indoor} & & & \\
Method & PSNR$\uparrow$ & SSIM$\uparrow$ & LPIPS$\downarrow$ & PSNR$\uparrow$ & SSIM$\uparrow$ & LPIPS$\downarrow$ & PSNR$\uparrow$ & SSIM$\uparrow$ & LPIPS$\downarrow$ & PSNR$\uparrow$ & SSIM$\uparrow$ & LPIPS$\downarrow$ & PSNR$\uparrow$ & SSIM$\uparrow$ & LPIPS$\downarrow$ \\
\midrule
2DGS & 25.85 & 0.778 & 0.259 & 30.51 & 0.929 & 0.169 & 27.03 & 0.835 & 0.166 & 31.19 & 0.945 & 0.093 & 23.66 & 0.832 & 0.280 \\
TriangleGS & 25.89 & \cellcolor{best}0.787 & \cellcolor{best}0.203 & 28.27 & 0.902 & 0.167 & 26.56 & 0.823 & 0.168 & 29.05 & 0.913 & 0.117 & 23.25 & 0.832 & \cellcolor{best}0.239 \\
DRK-GS (sparse) & 23.78 & 0.678 & 0.351 & 27.20 & 0.875 & 0.215 & 24.63 & 0.745 & 0.262 & 27.82 & 0.903 & 0.133 & $\times$ & $\times$ & $\times$ \\
DRK-GS (middle) & 24.12 & 0.703 & 0.320 & 27.47 & 0.882 & 0.207 & 24.94 & 0.762 & 0.238 & 27.78 & 0.903 & 0.131 & $\times$ & $\times$ & $\times$ \\
DRK-GS (dense) & 24.47 & 0.729 & 0.285 & 27.62 & 0.885 & 0.199 & 25.15 & 0.778 & 0.213 & 27.82 & 0.904 & 0.128 & 22.20 & 0.794 & 0.310 \\
Ours (PWL) & \cellcolor{second}25.93 & \cellcolor{second}0.784 & \cellcolor{second}0.243 & 30.85 & \cellcolor{best}0.933 & \cellcolor{best}0.160 & \cellcolor{second}27.08 & \cellcolor{second}0.838 & \cellcolor{best}0.157 & \cellcolor{second}31.36 & \cellcolor{second}0.947 & \cellcolor{best}0.088 & 23.86 & \cellcolor{second}0.835 & \cellcolor{second}0.272 \\
Ours (Pchip) & 25.89 & 0.782 & 0.246 & 30.87 & \cellcolor{best}0.933 & \cellcolor{second}0.161 & 27.06 & 0.837 & \cellcolor{second}0.159 & 31.30 & \cellcolor{second}0.947 & \cellcolor{second}0.089 & \cellcolor{second}23.91 & \cellcolor{second}0.835 & 0.274 \\
Ours (PWL-\textbf{sr}) & \cellcolor{best}25.94 & \cellcolor{second}0.784 & 0.245 & \cellcolor{best}30.89 & \cellcolor{best}0.933 & \cellcolor{best}0.160 & \cellcolor{best}27.11 & \cellcolor{best}0.839 & \cellcolor{best}0.157 & \cellcolor{best}31.43 & \cellcolor{best}0.948 & \cellcolor{best}0.088 & \cellcolor{best}23.92 & \cellcolor{best}0.836 & 0.273 \\
Ours (Pchip-\textbf{sr}) & 25.87 & 0.782 & 0.248 & \cellcolor{second}30.88 & \cellcolor{second}0.932 & \cellcolor{second}0.161 & 27.07 & \cellcolor{second}0.838 & \cellcolor{second}0.159 & 31.33 & 0.946 & \cellcolor{second}0.089 & \cellcolor{second}23.91 & \cellcolor{second}0.835 & 0.275 \\
\midrule
3DHGS$^\dagger$ & 26.40 & 0.798 & 0.175 & 32.54 & 0.948 & 0.093 & 27.83 & 0.857 & 0.111 & 33.29 & 0.962 & 0.046 & -- & -- & -- \\
BetaGS$^\dagger$ & 26.55 & 0.817 & 0.177 & 32.65 & 0.950 & 0.122 & 27.70 & 0.859 & 0.120 & 33.50 & 0.964 & 0.060 & 24.85$^*$ & 0.870$^*$ & 0.140$^*$ \\
\bottomrule
\end{tabular}%
}
\end{table}

As shown in ~\tref{tab:main_comparison}, our method consistently outperforms 2DGS across all dataset splits and metrics, with particularly notable gains on indoor scenes, and demonstrates more balanced performance across resolutions and scene types than TriangleGS. Against 3D methods (3DHGS, BetaGS), our approach narrows the gap, showing that our kernels can effectively enhance rendering quality within the 2D framework. Per-scene results and training time are provided in~\aref{app:per_scene}, and qualitative comparisons in~\aref{app:qual}.

\subsection{PSNR vs. \# of Gaussians}
\begin{figure}[htbp]
  \centering
  \includegraphics[width=1\textwidth]{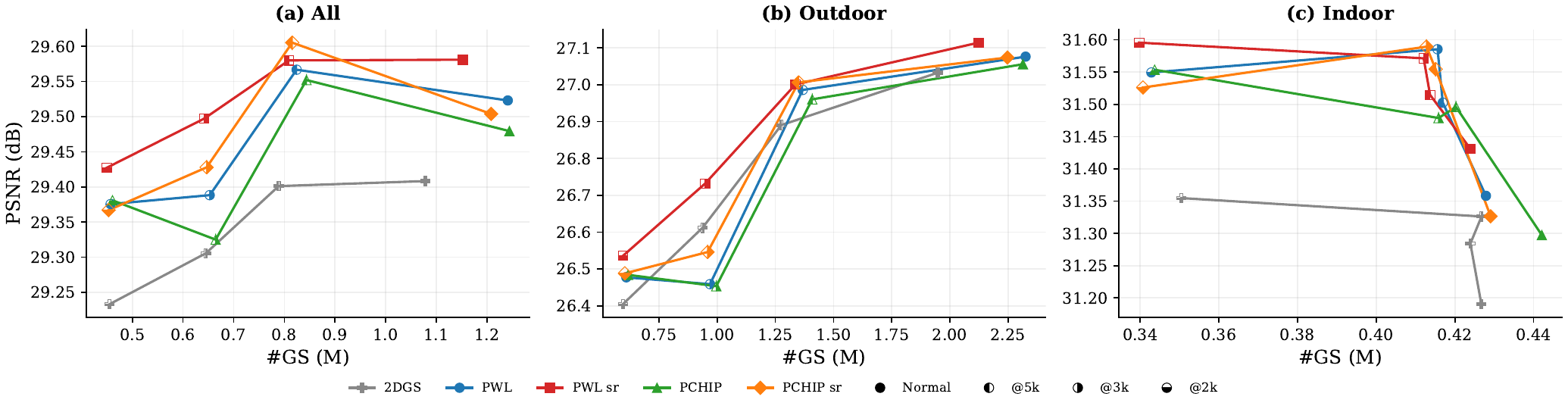}
   \caption{\textbf{PSNR vs. number of Gaussians on MipNeRF360 (×8)}. Markers denote densification schedules: Normal (15k iter), @5k, @3k, and @2k. Our methods achieve higher PSNR than 2DGS at comparable or fewer Gaussians. For indoor scenes (c), early densification termination improves quality, indicating that per-kernel expressiveness compensates for fewer primitives. Lines connect markers in schedule order; 2DGS indoor @3k exceeds @5k in count because pruning halts together with densification.}
   \label{fig:gaussian_number}
\end{figure}

To explore the quality–efficiency trade-off of our method, we vary the number of Gaussians by terminating densification early and compare against 2DGS under the same settings. In theory, the Gibbs phenomenon can be alleviated by increasing the number of basis functions, which in our case corresponds to using more Gaussian primitives. Therefore, evaluating this trade-off is crucial for highlighting the effectiveness of the proposed method. As shown in ~\fref{fig:gaussian_number}, our method consistently outperforms 2DGS across all Gaussian budgets. Notably, for indoor scenes, early termination of densification improves reconstruction quality, suggesting that the expressiveness of piecewise interpolation kernels effectively compensates for fewer primitives. Furthermore, the slope ratio (sr) parameterization provides additional gains in the low-Gaussian regime, consistent with its broader coverage in the control point space.

Our results also reveal that Pchip, despite offering a smoother deformation function, does not yield a notable improvement over PWL. This suggests that Gaussian reconstruction benefits more from the ability to represent asymmetric and high-frequency opacity variations than from smoothness. The slope-ratio parameterization improves PWL in PSNR across all splits, owing to its broader exploration of the control-point parameter space, which enables more diverse opacity shapes per primitive. For Pchip, however, it brings no consistent gain.

\subsection{Geometry Evaluation}
\label{sec:geometry}

\begin{table}[t]
\centering
\small
\setlength{\tabcolsep}{2.4pt}
\begin{minipage}[t]{0.49\linewidth}
\centering
\caption{\textbf{F1 score on Tanks \& Temples.} Meetingroom is excluded for
all methods.}
\label{tab:tnt_f1}
\vspace{3pt}
\begin{tabular}{lcccccc}
\toprule
Method & Barn & Cat. & Court. & Ign. & Truck & Mean \\
\midrule
2DGS & \cellcolor{best}0.386 & \cellcolor{second}0.248 & \cellcolor{second}0.120 & \cellcolor{best}0.496 & 0.465 & \cellcolor{second}0.343 \\
PWL & 0.368 & 0.235 & 0.106 & 0.438 & 0.506 & 0.331 \\
PWL-\textbf{sr} & 0.380 & \cellcolor{best}0.249 & \cellcolor{best}0.131 & \cellcolor{second}0.469 & \cellcolor{second}0.508 & \cellcolor{best}0.347 \\
Pchip & 0.376 & 0.230 & 0.104 & 0.433 & \cellcolor{best}0.533 & 0.335 \\
Pchip-\textbf{sr} & \cellcolor{second}0.383 & 0.240 & 0.056 & 0.404 & 0.505 & 0.318 \\
\bottomrule
\end{tabular}
\end{minipage}\hfill
\begin{minipage}[t]{0.49\linewidth}
\centering
\caption{\textbf{Boundary PSNR on Tanks \& Temples.} Ours as differences
to 2DGS.}
\label{tab:boundary}
\vspace{3pt}
\begin{tabular}{lcccccc}
\toprule
Method & $r{=}1$ & $r{=}2$ & $r{=}4$ & $r{=}8$ & $r{=}16$ & full \\
\midrule
2DGS & 18.53 & 19.32 & 20.20 & 21.12 & 21.96 & 23.32 \\
PWL & +.130 & +.151 & +.173 & +.190 & +.211 & +.239 \\
PWL-\textbf{sr} & \cellcolor{best}+.209 & \cellcolor{best}+.238 & \cellcolor{best}+.267 & \cellcolor{best}+.294 & \cellcolor{best}+.318 & \cellcolor{best}+.356 \\
Pchip & \cellcolor{second}+.133 & \cellcolor{second}+.161 & \cellcolor{second}+.187 & \cellcolor{second}+.208 & \cellcolor{second}+.225 & \cellcolor{second}+.278 \\
Pchip-\textbf{sr} & +.034 & +.035 & +.035 & +.045 & +.067 & +.107 \\
\bottomrule
\end{tabular}
\end{minipage}
\end{table}

Sharper opacity kernels do not come at the cost of geometry. ~\tref{tab:tnt_f1} shows that PWL-sr matches 2DGS in mean F1, and that all four variants improve on Truck while dropping on Ignatius. Truck is dominated by planar panels and sharp man-made edges, where an asymmetric falloff helps, whereas the smooth curved surfaces of Ignatius are already well served by a symmetric Gaussian. We follow the 2DGS protocol and exclude Meetingroom for all methods because it exceeds our 64\,GB host memory. The low Pchip-sr score on Courthouse comes from the ICP registration step of the official evaluation script, and we report it unchanged.

\subsection{Boundary Evaluation}
\label{sec:boundary}
\begin{figure}[h]
  \centering
  \setlength{\tabcolsep}{1pt}
  \begin{tabular}{cccc}
    \includegraphics[width=0.24\linewidth]{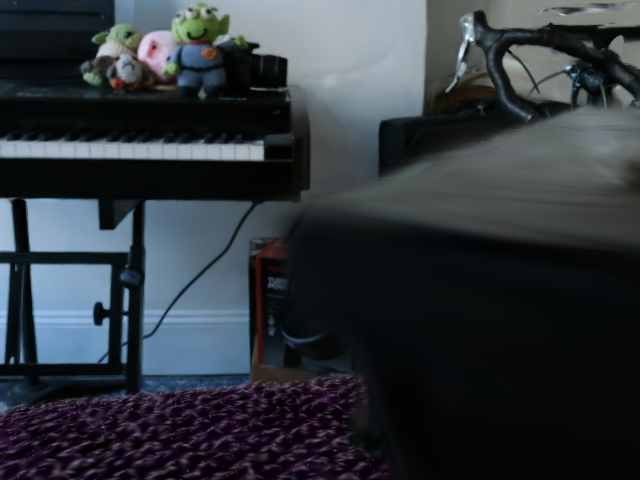} &
    \includegraphics[width=0.24\linewidth]{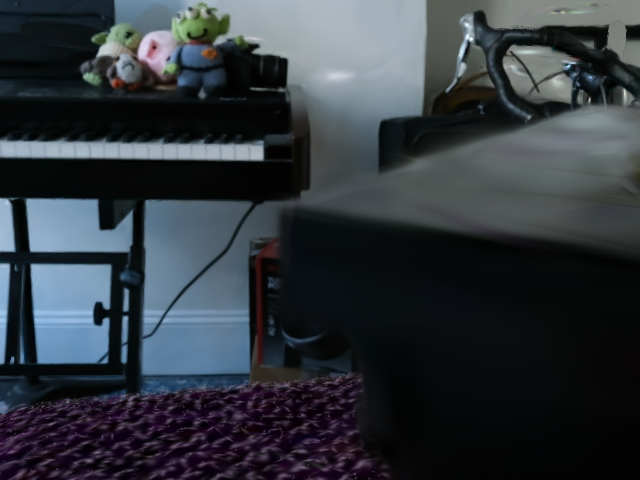} &
    \includegraphics[width=0.24\linewidth]{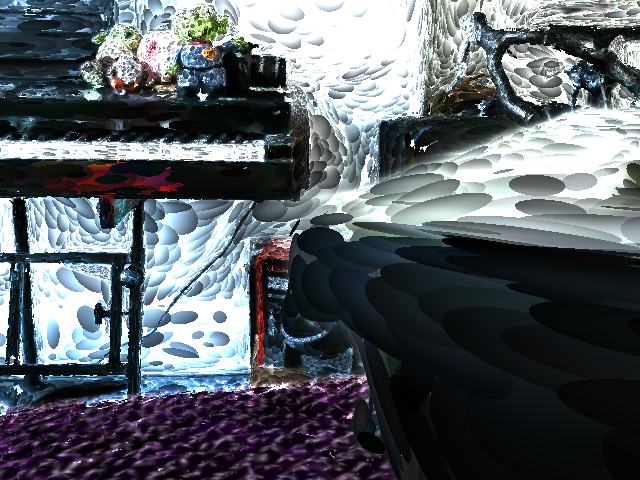} &
    \includegraphics[width=0.24\linewidth]{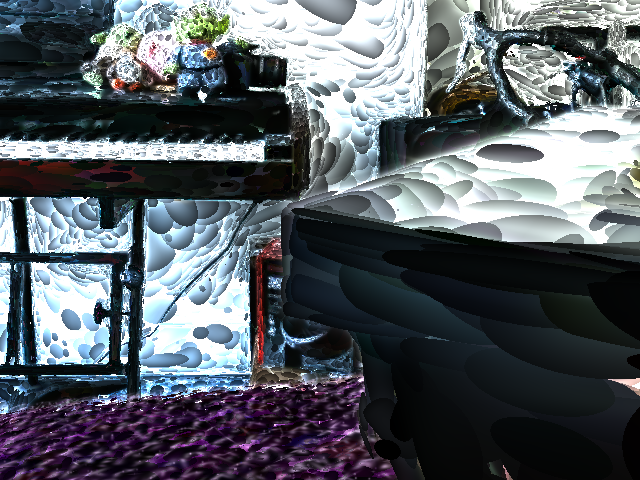} \\
    {\small 2DGS RGB} & {\small PWL \textbf{sr} RGB (Ours)} & {\small 2DGS Ellipse} & {\small PWL \textbf{sr} Ellipse (Ours)}
  \end{tabular}
  \caption{\textbf{Rendered images and Gaussian ellipse visualizations.} Our warped Gaussians align more tightly with object boundaries compared to the symmetric ellipses of 2DGS.}
  \label{fig:ellipse}
\end{figure}

The gains of our kernel are not confined to smooth regions. Following the band protocol of Boundary IoU~\cite{cheng2021boundary}, we compute PSNR within bands around ground-truth Sobel edges dilated by radius $r$. ~\tref{tab:boundary} shows that every variant improves boundary PSNR over 2DGS at every radius, with PWL-sr remaining the strongest, and SSIM results are given in~\aref{app:boundary}. The gain in the narrowest bands is  smaller than over the full image, since a boundary pixel is composited from many overlapping splats and alpha blending partly averages out the effect of sharpening any single kernel. ~\fref{fig:ellipse} visualizes the corresponding Gaussian ellipses. At object boundaries, 2DGS stacks many overlapping symmetric ellipses and blurs sharp edges, whereas our warped kernels align with boundary contours and produce cleaner edges.

\section{Discussion}
\label{sec:discussion}

\textbf{Limitations.}
Our warp uses a single interior knot per axis. In the preliminary multi-knot experiments reported in~\aref{app:multiknot}, adjacent knots tended to drift together and, once they coincided, failed to re-separate, a known degeneracy of free-knot splines~\cite{jupp1978Approximation,beliakov2004Least}. Progressive knot insertion is a promising remedy that we leave to future work. The axis-separated warp also cannot place a sharp transition along edges not aligned with the tangent axes, so residual ringing may persist at acute structures. Extending the kernel to 3DGS is tractable, since the PWL warp keeps the kernel Gaussian between consecutive knot planes, with the half-space integral of 3D-HGS~\cite{li20253DHGS} as the single-plane case. For dynamic scenes, existing 2DGS methods~\cite{wang2024Spacetime,matsuki20254DTAM} deform surfels with rotations only, and how warped kernels behave under non-rigid deformation remains unexplored. Finally, training takes roughly 35--45\% longer than 2DGS, which tight bounding-box techniques such as Speedy-Splat~\cite{hanson2025SpeedySplat} and FlashGS~\cite{feng2025FlashGS} could partly recover.

\textbf{Conclusion.} We present a learnable warped kernel for 2DGS that replaces the symmetric Gaussian opacity profile with axis-separated piecewise linear profiles. This design introduces asymmetric falloff while preserving differentiability, enabling sharp boundaries that symmetric smooth kernels cannot represent. Experiments on MipNeRF360 and Tanks\,\&\,Temples show the best PSNR among 2DGS-based approaches with a comparable number of Gaussians.
\newpage

\begin{ack}
This research was supported by the Government of Ontario through the Early Researcher
Awards program (Award No. ER24-18-220). The authors declare no competing interests.
\end{ack}
\small \bibliographystyle{plainnat-mine} \bibliography{references, main}

\newpage
\appendix

\section{Square Wave Fitting}
\label{app:algorithm}

To systematically evaluate each kernel's ability to represent sharp features, we design a greedy incremental fitting procedure that iteratively adds one kernel component at a time, initializing each at the dominant residual peak and refitting all parameters jointly via Levenberg--Marquardt optimization~\cite{roweis1996levenberg}. The Bayesian Information Criterion (BIC) governs model selection, balancing reconstruction accuracy against model complexity. The full procedure is summarized in Algorithm~\ref{alg:greedy-gaussian}.

\begin{algorithm}[htbp]
\caption{Greedy Incremental Gaussian Mixture Fitting}
\label{alg:greedy-gaussian}
\begin{algorithmic}[1]
\Require Spectrum $(x, y)$, kernel class $\mathcal{K}$, max components $N_{\max}$, BIC threshold $\delta$
\Ensure Optimal number of components $n^*$, fitted parameters $\boldsymbol{\theta}^*$

\State $r \gets y$ \Comment{Initialize residual}
\State $\mathrm{BIC}^* \gets +\infty$, \; $n^* \gets 0$, \; $\boldsymbol{\theta}^* \gets \varnothing$
\State $\mathcal{C} \gets \varnothing$ \Comment{Set of kernel components}

\For{$n = 1, 2, \ldots, N_{\max}$}
    \State Detect peaks $\mathcal{P} \gets \textsc{FindPeaks}(r)$ with height $\geq 0.1 \cdot \max(r)$
    \If{$\mathcal{P} = \varnothing$}
        \State $\mathcal{P} \gets \{\arg\max_i \, r_i\}$
    \EndIf
    \State $j^* \gets \arg\max_{j \in \mathcal{P}} \, r_j$ \Comment{Select dominant residual peak}
    \State Initialize new kernel: $\mu \gets x_{j^*}$, \; $a \gets r_{j^*}$, \; $\sigma \gets \frac{x_{\max} - x_{\min}}{4n}$
    \State $\mathcal{C} \gets \mathcal{C} \cup \{\mathcal{K}(\mu, \sigma, a)\}$
    \State $\boldsymbol{\theta}_0 \gets$ concatenate parameters from all kernels in $\mathcal{C}$
    \State $\boldsymbol{\theta} \gets \textsc{LevenbergMarquardt}\!\left(\min_{\boldsymbol{\theta}} \left\| y - \sum_{k \in \mathcal{C}} k(x; \boldsymbol{\theta}) \right\|^2, \; \boldsymbol{\theta}_0\right)$
    \State $\hat{y} \gets \sum_{k \in \mathcal{C}} k(x; \boldsymbol{\theta})$
    \State $\mathrm{RSS} \gets \sum_i (y_i - \hat{y}_i)^2$
    \State $\mathrm{BIC} \gets N \ln\!\left(\frac{\mathrm{RSS}}{N}\right) + |\boldsymbol{\theta}| \ln N$ \Comment{Bayesian Information Criterion}
    \State $r \gets y - \hat{y}$ \Comment{Update residual}
    \If{$\mathrm{BIC} < \mathrm{BIC}^* - \delta$} \Comment{Accept only if BIC improves}
        \State $\mathrm{BIC}^* \gets \mathrm{BIC}$, \; $\boldsymbol{\theta}^* \gets \boldsymbol{\theta}$, \; $n^* \gets n$
    \EndIf
\EndFor

\State \Return $n^*$, $\boldsymbol{\theta}^*$
\end{algorithmic}
\end{algorithm}

\begin{figure}[htbp]
  \centering
  \includegraphics[width=1\textwidth]{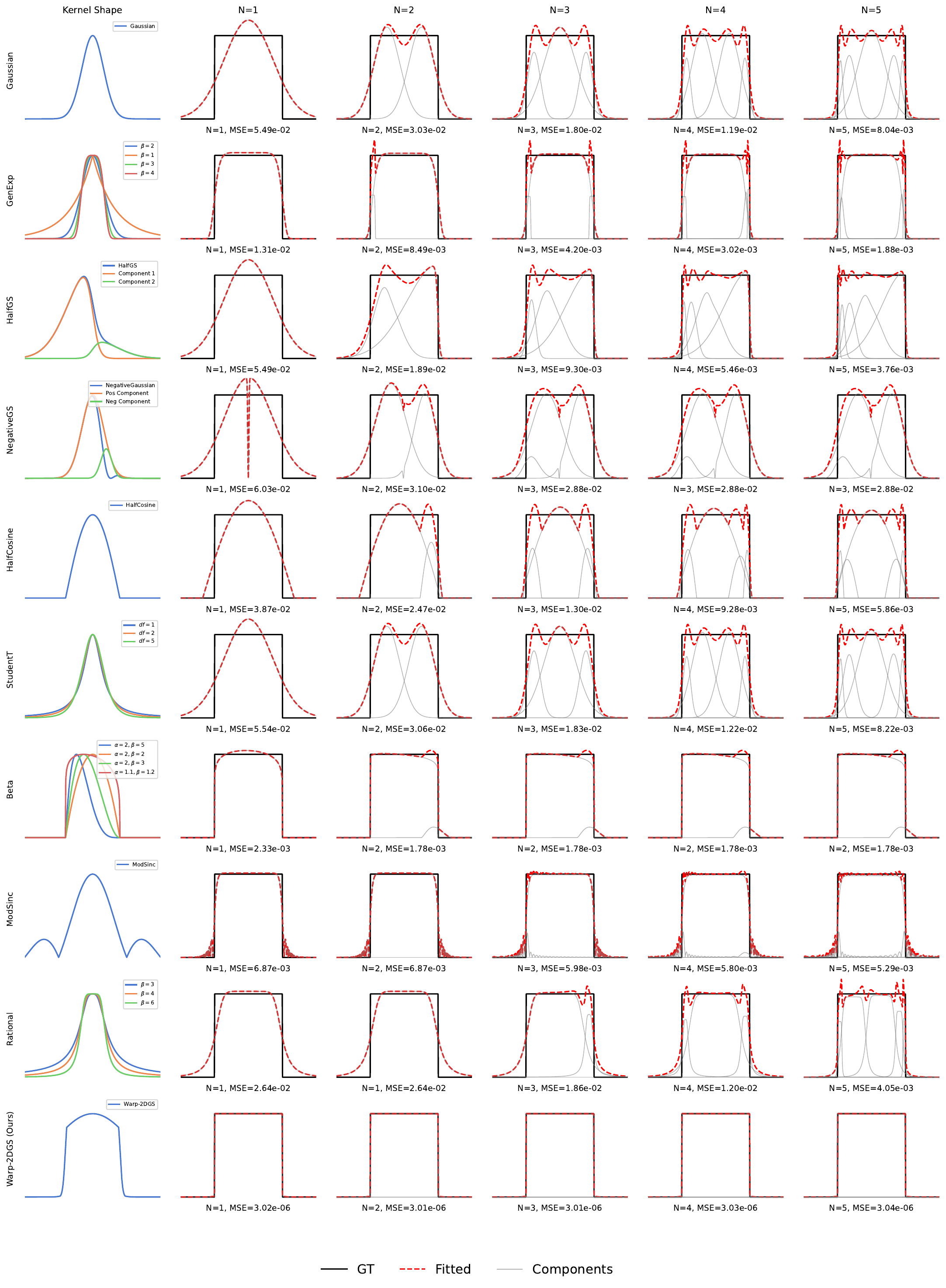}
   \caption{\textbf{Qualitative comparison of kernel functions for 1D square wave signal fitting. } Each row corresponds to a different kernel type. The first column illustrates the kernel shape under representative parameterizations. The remaining columns show the fitting results on a square wave target signal using an increasing number of components ($N=1,2,3,4,5$), where the black curve denotes the ground truth, the red dashed curve denotes the overall fit, and gray curves denote individual components. Notably, our warped Gaussian kernel exhibits nearly no Gibbs-like ringing artifacts near discontinuities, even with a small number of components. Our kernel uses the same center-symmetric variant as in~\fref{fig:gibbs}.}
   \label{fig:gaussian_kernals_comparsion}
\end{figure}

~\fref{fig:gaussian_kernals_comparsion} presents the results for each kernel under identical conditions. As the number of components increases, symmetric kernels consistently exhibit Gibbs-like oscillations near the discontinuities at $x=0.25$ and $x=0.75$. In contrast, our proposed warped Gaussian kernel suppresses these ringing artifacts even with few components, as its asymmetric profiles can conform to sharp transitions without compensatory oscillation from neighboring primitives.

\section{Per-Scene Quantitative Results}
\label{app:per_scene}

~\tref{tab:mipnerf360_detail} reports per-scene results on MipNeRF 360 at $\times 4$ and $\times 8$ resolutions. Our variants achieve the best or second-best scores among 2D methods across nearly all scenes and metrics. The training time of our method lies between 2DGS and the other kernel-based methods, and remains substantially lower than that of DRK-GS and TriangleGS.

\begin{table}[ht]
\centering
\caption{\textbf{Per-scene results on MipNeRF 360.} \colorbox{best}{Best} and \colorbox{second}{second} among 2D methods. \textbf{sr}: slope parameterization method. $^\dagger$: 3D methods. Train: average training time (min).}
\label{tab:mipnerf360_detail}
\setlength{\tabcolsep}{1.5pt}
\resizebox{\textwidth}{!}{
\tiny
\begin{tabular}{l|ccccccc|ccccccc|ccccccc|c}
\toprule
& \multicolumn{7}{c|}{PSNR$\uparrow$} & \multicolumn{7}{c|}{SSIM$\uparrow$} & \multicolumn{7}{c|}{LPIPS$\downarrow$} & \\
Method & Bicy. & Bons. & Coun. & Gard. & Kitc. & Room & Stump & Bicy. & Bons. & Coun. & Gard. & Kitc. & Room & Stump & Bicy. & Bons. & Coun. & Gard. & Kitc. & Room & Stump & Train \\
\midrule
\multicolumn{23}{c}{\textit{$\times 4$ resolution}} \\
\midrule
2DGS & 24.75 & 31.65 & 28.55 & 26.68 & 30.38 & 31.48 & 26.11 & 0.734 & 0.945 & 0.906 & 0.843 & 0.930 & 0.933 & 0.756 & 0.305 & 0.170 & 0.185 & 0.168 & 0.114 & 0.206 & 0.303 & 13.6 \\
TriangleGS & 24.56 & 28.15 & 27.58 & \cellcolor{best}27.10 & 28.75 & 28.59 & 26.01 & \cellcolor{best}0.752 & 0.897 & 0.895 & \cellcolor{best}0.860 & 0.915 & 0.903 & 0.750 & \cellcolor{best}0.227 & 0.187 & \cellcolor{best}0.165 & \cellcolor{best}0.123 & 0.128 & \cellcolor{best}0.186 & \cellcolor{best}0.257 & 29.9 \\
DRK-GS (sparse) & 23.15 & 27.87 & 26.60 & 23.97 & 26.03 & 28.30 & 24.23 & 0.631 & 0.888 & 0.864 & 0.744 & 0.855 & 0.894 & 0.660 & 0.388 & 0.217 & 0.223 & 0.263 & 0.172 & 0.250 & 0.401 & 68.7 \\
DRK-GS (middle) & 23.43 & 28.36 & 26.76 & 24.41 & 26.43 & 28.33 & 24.51 & 0.660 & 0.902 & 0.868 & 0.766 & 0.864 & 0.895 & 0.684 & 0.355 & 0.201 & 0.217 & 0.237 & 0.164 & 0.244 & 0.370 & 74.9 \\
DRK-GS (dense) & 23.77 & 28.15 & 26.96 & 24.99 & 26.75 & 28.65 & 24.63 & 0.691 & 0.895 & 0.873 & 0.790 & 0.869 & 0.902 & 0.707 & 0.316 & 0.197 & 0.208 & 0.205 & 0.159 & 0.233 & 0.334 & 90.4 \\
Ours (PWL) & 24.74 & 32.04 & 28.73 & 26.80 & \cellcolor{best}31.23 & 31.41 & \cellcolor{best}26.24 & 0.741 & \cellcolor{best}0.949 & \cellcolor{best}0.910 & \cellcolor{second}0.848 & \cellcolor{best}0.939 & \cellcolor{second}0.934 & \cellcolor{best}0.763 & \cellcolor{second}0.286 & \cellcolor{best}0.159 & \cellcolor{second}0.177 & \cellcolor{second}0.155 & \cellcolor{best}0.103 & \cellcolor{second}0.199 & \cellcolor{second}0.288 & 19.6 \\
Ours (Pchip) & \cellcolor{best}24.77 & 32.04 & \cellcolor{best}28.75 & 26.77 & \cellcolor{second}31.19 & \cellcolor{second}31.51 & 26.13 & 0.741 & \cellcolor{second}0.948 & \cellcolor{second}0.909 & 0.846 & \cellcolor{best}0.939 & \cellcolor{second}0.934 & 0.760 & 0.289 & 0.162 & 0.179 & 0.157 & \cellcolor{best}0.103 & 0.201 & 0.292 & 21.9 \\
Ours (PWL-\textbf{sr}) & \cellcolor{best}24.77 & \cellcolor{second}32.08 & \cellcolor{second}28.74 & \cellcolor{second}26.84 & 31.18 & \cellcolor{best}31.57 & \cellcolor{second}26.20 & \cellcolor{second}0.742 & \cellcolor{best}0.949 & \cellcolor{best}0.910 & \cellcolor{second}0.848 & \cellcolor{best}0.939 & \cellcolor{best}0.935 & \cellcolor{best}0.763 & 0.289 & \cellcolor{second}0.160 & \cellcolor{second}0.177 & 0.156 & \cellcolor{best}0.103 & \cellcolor{second}0.199 & 0.290 & 19.9 \\
Ours (Pchip-\textbf{sr}) & \cellcolor{second}24.76 & \cellcolor{best}32.11 & \cellcolor{second}28.74 & 26.71 & 31.17 & \cellcolor{second}31.51 & 26.15 & 0.740 & \cellcolor{best}0.949 & \cellcolor{second}0.909 & 0.846 & \cellcolor{second}0.938 & \cellcolor{second}0.934 & \cellcolor{second}0.761 & 0.292 & 0.161 & 0.179 & 0.159 & \cellcolor{second}0.104 & 0.202 & 0.292 & 21.8 \\
\midrule
3DHGS$^\dagger$ & 25.28 & 33.98 & 30.15 & 27.56 & 33.09 & 32.93 & 26.35 & 0.767 & 0.962 & 0.928 & 0.867 & 0.952 & 0.949 & 0.759 & 0.200 & 0.093 & 0.106 & 0.105 & 0.063 & 0.108 & 0.219 & 19.3 \\
BetaGS$^\dagger$ & 25.28 & 34.18 & 30.50 & 27.58 & 33.05 & 32.89 & 26.80 & 0.784 & 0.966 & 0.935 & 0.870 & 0.952 & 0.949 & 0.796 & 0.198 & 0.118 & 0.131 & 0.116 & 0.085 & 0.154 & 0.216 & 23.2 \\
\midrule
\multicolumn{23}{c}{\textit{$\times 8$ resolution}} \\
\midrule
2DGS & \cellcolor{second}26.20 & 32.02 & 29.03 & 28.25 & 31.60 & \cellcolor{second}32.11 & 26.65 & 0.814 & \cellcolor{second}0.956 & 0.921 & 0.897 & \cellcolor{second}0.955 & \cellcolor{second}0.950 & 0.795 & 0.189 & 0.088 & \cellcolor{second}0.111 & 0.095 & 0.059 & 0.113 & 0.215 & 6.6 \\
TriangleGS & 25.90 & 28.66 & 28.28 & 27.30 & 30.03 & 29.23 & 26.48 & 0.798 & 0.900 & 0.906 & 0.884 & 0.933 & 0.913 & 0.788 & 0.189 & 0.133 & 0.117 & 0.103 & 0.085 & 0.134 & 0.212 & 15.1 \\
DRK-GS (sparse) & 24.38 & 28.67 & 26.97 & 25.12 & 26.98 & 28.65 & 24.41 & 0.723 & 0.917 & 0.886 & 0.816 & 0.896 & 0.914 & 0.695 & 0.282 & 0.122 & 0.145 & 0.172 & 0.108 & 0.156 & 0.332 & 35.2 \\
DRK-GS (middle) & 24.61 & 28.65 & 27.06 & 25.43 & 26.73 & 28.68 & 24.77 & 0.743 & 0.917 & 0.888 & 0.828 & 0.889 & 0.915 & 0.716 & 0.255 & 0.120 & 0.142 & 0.157 & 0.111 & 0.152 & 0.302 & 37.5 \\
DRK-GS (dense) & 24.84 & 28.31 & 27.15 & 25.89 & 27.07 & 28.76 & 24.73 & 0.762 & 0.908 & 0.891 & 0.844 & 0.897 & 0.919 & 0.729 & 0.224 & 0.121 & 0.138 & 0.138 & 0.105 & 0.147 & 0.278 & 45.2 \\
Ours (PWL) & 26.18 & \cellcolor{best}32.41 & \cellcolor{second}29.27 & 28.34 & \cellcolor{second}31.88 & 31.87 & \cellcolor{best}26.71 & \cellcolor{second}0.817 & \cellcolor{best}0.958 & \cellcolor{second}0.924 & \cellcolor{second}0.899 & \cellcolor{best}0.956 & 0.949 & \cellcolor{best}0.798 & \cellcolor{best}0.177 & \cellcolor{best}0.082 & \cellcolor{best}0.104 & \cellcolor{best}0.090 & \cellcolor{best}0.056 & \cellcolor{second}0.112 & \cellcolor{best}0.205 & 8.9 \\
Ours (Pchip) & 26.14 & \cellcolor{second}32.38 & 29.23 & \cellcolor{second}28.35 & 31.84 & 31.74 & 26.67 & 0.816 & \cellcolor{best}0.958 & \cellcolor{second}0.924 & 0.898 & \cellcolor{best}0.956 & 0.948 & \cellcolor{second}0.797 & \cellcolor{second}0.178 & \cellcolor{second}0.083 & \cellcolor{best}0.104 & \cellcolor{second}0.092 & \cellcolor{best}0.056 & 0.113 & \cellcolor{second}0.206 & 9.9 \\
Ours (PWL-\textbf{sr}) & \cellcolor{best}26.21 & \cellcolor{second}32.38 & \cellcolor{second}29.27 & \cellcolor{best}28.43 & \cellcolor{best}31.90 & \cellcolor{best}32.17 & \cellcolor{second}26.70 & \cellcolor{best}0.818 & \cellcolor{best}0.958 & \cellcolor{best}0.925 & \cellcolor{best}0.900 & \cellcolor{best}0.956 & \cellcolor{best}0.952 & \cellcolor{best}0.798 & \cellcolor{best}0.177 & \cellcolor{best}0.082 & \cellcolor{best}0.104 & \cellcolor{best}0.090 & \cellcolor{best}0.056 & \cellcolor{best}0.108 & \cellcolor{best}0.205 & 8.9 \\
Ours (Pchip-\textbf{sr}) & 26.19 & \cellcolor{best}32.41 & \cellcolor{best}29.28 & 28.33 & 31.67 & 31.94 & \cellcolor{second}26.70 & \cellcolor{second}0.817 & \cellcolor{best}0.958 & \cellcolor{best}0.925 & \cellcolor{second}0.899 & \cellcolor{best}0.956 & 0.948 & \cellcolor{best}0.798 & 0.180 & \cellcolor{second}0.083 & \cellcolor{best}0.104 & \cellcolor{second}0.092 & \cellcolor{second}0.057 & 0.114 & \cellcolor{second}0.206 & 10.2 \\
\midrule
3DHGS$^\dagger$ & 26.88 & 34.45 & 30.73 & 29.61 & 34.32 & 33.67 & 27.00 & 0.846 & 0.972 & 0.942 & 0.922 & 0.971 & 0.965 & 0.804 & 0.120 & 0.041 & 0.062 & 0.057 & 0.030 & 0.051 & 0.156 & 11.9 \\
BetaGS$^\dagger$ & 26.70 & 34.65 & 30.96 & 29.27 & 34.42 & 33.96 & 27.14 & 0.841 & 0.973 & 0.945 & 0.914 & 0.971 & 0.968 & 0.821 & 0.126 & 0.052 & 0.072 & 0.068 & 0.039 & 0.076 & 0.167 & 19.6 \\
\bottomrule
\end{tabular}
}
\end{table}

\section{Boundary Evaluation}
\label{app:boundary}

~\tref{tab:boundary_ssim} complements the boundary PSNR in
~\tref{tab:boundary} with SSIM computed over the same bands. PWL and PWL-sr improve boundary SSIM at every radius, whereas Pchip and Pchip-sr fall slightly below 2DGS.

\begin{table}[ht]
\centering
\small
\setlength{\tabcolsep}{4pt}
\caption{\textbf{Boundary SSIM on Tanks \& Temples.} The 2DGS row is absolute,
and the other rows are differences to 2DGS.}
\label{tab:boundary_ssim}
\vspace{3pt}
\begin{tabular}{lcccccc}
\toprule
Method & $r{=}1$ & $r{=}2$ & $r{=}4$ & $r{=}8$ & $r{=}16$ & full \\
\midrule
2DGS & 0.7538 & 0.7462 & 0.7447 & 0.7591 & 0.7785 & 0.8116 \\
PWL & \cellcolor{second}+0.0029 & \cellcolor{second}+0.0029 & \cellcolor{second}+0.0027 & \cellcolor{second}+0.0022 & \cellcolor{second}+0.0020 & \cellcolor{second}+0.0022 \\
PWL-\textbf{sr} & \cellcolor{best}+0.0067 & \cellcolor{best}+0.0064 & \cellcolor{best}+0.0057 & \cellcolor{best}+0.0050 & \cellcolor{best}+0.0045 & \cellcolor{best}+0.0043 \\
Pchip & $-$0.0005 & $-$0.0006 & $-$0.0010 & $-$0.0015 & $-$0.0017 & $-$0.0011 \\
Pchip-\textbf{sr} & $-$0.0037 & $-$0.0034 & $-$0.0030 & $-$0.0027 & $-$0.0021 & $-$0.0009 \\
\bottomrule
\end{tabular}
\end{table}

\section{Multi-Knot Variants}
\label{app:multiknot}

\begin{figure}[ht]
\centering
\includegraphics[width=\linewidth]{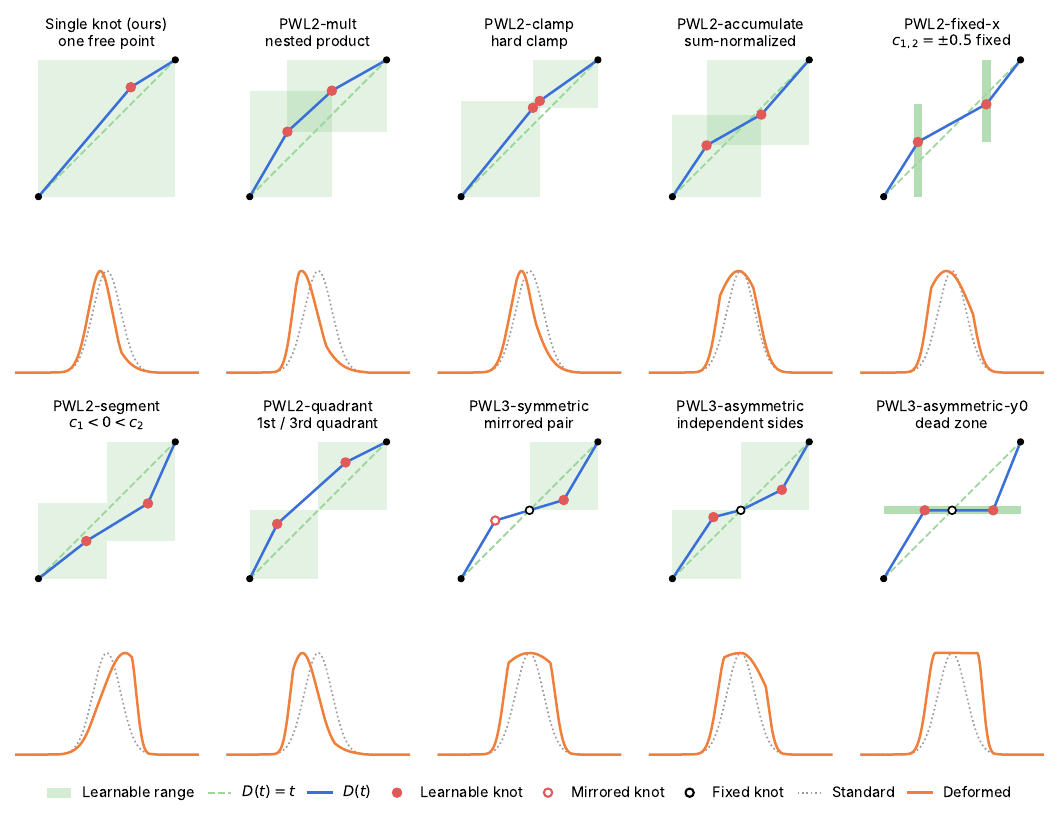}
\caption{\textbf{Multi-knot deformation variants.} For each variant, the top row shows the deformation function $D(t)$ and the bottom row shows the resulting kernel $G(r \cdot D(x/r))$. Green shading marks the range in which each learnable knot can move while the other knots are held fixed. Parameter values are chosen for illustration and are not learned values.}
\label{fig:multiknot}
\end{figure}

In early exploration, we tested a range of multi-knot parameterizations for both PWL and Pchip, shown in~\fref{fig:multiknot}. Monotonicity requires the interior knots to stay strictly ordered, so each variant needs its own way of enforcing this ordering. PWL2 adds a second learnable knot, and its variants differ only in how the ordering is enforced. PWL3 places a fixed knot at the origin with one learnable knot on each side.

These variants confound knot count with parameterization, and several were run under an earlier version of our pipeline, so we report qualitative trends rather than a controlled ablation. Subject to these caveats, none of the variants offered a clear advantage over the single-knot design. In PWL3, the two learnable knots lie on either side of a fixed origin and cannot collide with each other. PWL3-symmetric mirrors a single $(c_1, d_1)$ pair, so the added knots neither split the gradient signal across more parameters nor allow two knots to collide. It therefore isolates the effect of the extra segments alone, and it performs on par with the single-knot design. PWL3-asymmetric and PWL3-asymmetric-y0 have more free parameters and improve over the 2DGS baseline, but fall short of the single-knot design. In the PWL2 family, two free interior knots lie in the same interval, and we observed adjacent knots drifting together and, once they coincided, failing to re-separate, with unstable gradients as they approached each other. The last two rows of ~\tref{tab:sensitivity} report PWL2-mult and PWL2-segment under the current pipeline.
 
This failure mode is well documented in the free-knot spline literature. Merged knots form degenerate faces of the feasible region on which the gradient separating them vanishes identically~\cite{jupp1978Approximation}, and the number of such faces grows with the knot count. Local optimizers are reported to converge to these degenerate solutions rather than to the true optimum, particularly from uniformly spaced initializations~\cite{beliakov2004Least}. A viable remedy is progressive knot insertion, which trains to convergence with one knot and then inserts additional knots along the learned curve. For PWL kernels this insertion is an exact reparameterization that leaves the curve unchanged. Adopting it introduces further design choices, such as when to insert new knots and where to place them, and we leave a systematic investigation to future work.

\section{Rescale Factor}
 
The rescale factor $r$ sets the domain $G(-r, r)$ over which the control points can reshape the Gaussian. At $r = 1$ only the region near the peak is deformable, and a larger $r$ extends the deformable region into the tail. Since the deformation function satisfies $D(t) = t$ for $|t| > 1$ (Eq.~6), $r$ controls the size of the deformable region relative to the Gaussian, whose own extent is set independently by the scaling parameters. Too large an $r$ risks introducing invalid area, while too small an $r$ leaves insufficient room for deformation. As $r \to 0$, nearly all $t/r$ exceed 1 and the kernel degenerates back to a standard Gaussian. We test $r \in \{2, 3, 4\}$ for both the direct and slope-ratio parameterizations of PWL on MipNeRF 360 ($\times 8$). As shown in~\tref{tab:sensitivity}, neither $r = 2$ nor $r = 4$ significantly outperforms our default $r = 3$, in either outdoor or indoor scenes.

\begin{table}[ht]
\centering
\small
\setlength{\tabcolsep}{4pt}
\caption{\textbf{Sensitivity to the rescale factor and the number of knots on
MipNeRF 360 ($\times 8$).} $Q_{max}=3$ for all rows. The one-knot rows with
$r=3$ are the configurations used in the main paper.}
\label{tab:sensitivity}
\vspace{3pt}
\begin{tabular}{cccc@{\hspace{8pt}}ccc@{\hspace{8pt}}ccc}
\toprule
& & & & \multicolumn{3}{c}{Outdoor} & \multicolumn{3}{c}{Indoor} \\
\cmidrule(lr){5-7} \cmidrule(lr){8-10}
Kernel & Knots & $r$ & Param. & PSNR$\uparrow$ & SSIM$\uparrow$ & LPIPS$\downarrow$ & PSNR$\uparrow$ & SSIM$\uparrow$ & LPIPS$\downarrow$ \\
\midrule
PWL & 1 & 3 & Direct & \cellcolor{second}27.08 & \cellcolor{second}0.838 & 0.157 & 31.36 & \cellcolor{second}0.947 & \cellcolor{best}0.088 \\
PWL & 1 & 2 & Direct & \cellcolor{second}27.08 & \cellcolor{best}0.839 & \cellcolor{best}0.155 & 31.23 & 0.946 & \cellcolor{best}0.088 \\
PWL & 1 & 4 & Direct & 27.03 & 0.836 & 0.161 & 31.38 & \cellcolor{second}0.947 & \cellcolor{second}0.089 \\
\midrule
PWL & 1 & 3 & \textbf{sr} & \cellcolor{best}27.11 & \cellcolor{best}0.839 & 0.157 & \cellcolor{best}31.43 & \cellcolor{best}0.948 & \cellcolor{best}0.088 \\
PWL & 1 & 2 & \textbf{sr} & \cellcolor{second}27.08 & \cellcolor{best}0.839 & \cellcolor{second}0.156 & 31.30 & \cellcolor{second}0.947 & \cellcolor{second}0.089 \\
PWL & 1 & 4 & \textbf{sr} & 27.03 & 0.836 & 0.161 & \cellcolor{second}31.39 & \cellcolor{second}0.947 & \cellcolor{second}0.089 \\
\midrule
PWL & 2 & 3 & mult & 26.99 & 0.836 & \cellcolor{second}0.156 & 30.94 & 0.943 & 0.090 \\
PWL & 2 & 3 & segment & 26.96 & 0.836 & 0.158 & 31.14 & 0.945 & 0.090 \\
\bottomrule
\end{tabular}
\end{table}

\section{Qualitative Comparison}
\label{app:qual}

~\fref{fig:qualitative} shows qualitative comparisons on selected scenes from MipNeRF 360. TriangleGS and DRK-GS tend to produce blurry or noisy textures in fine-detail regions, while 2DGS preserves overall structure but over-smooths sharp edges. Our method recovers sharper boundaries and finer textures, consistent with the quantitative improvements reported in ~\tref{tab:mipnerf360_detail}.


\newcommand{\imgdir}{main/fig/appendix/qualitative}

\newcommand{\frameA}{bonsai_4_DSCF5597.png}
\newcommand{\frameB}{bonsai_4_DSCF5637.png}
\newcommand{\frameC}{counter_4_DSCF6089.png}
\newcommand{\frameD}{stump_4__DSC9237.png}
\newcommand{\frameE}{room_4_DSCF4971.png}

\newcommand{\mGT}{GT}
\newcommand{\mTriangle}{Triangle}
\newcommand{\mDRK}{DRK}
\newcommand{\mTwoDGS}{2DGS}
\newcommand{\mOurs}{Ours_PWL}

\newcommand{\qualimg}[4]{%
  \begin{tikzpicture}[spy using outlines={rectangle, red, dashed, magnification=#4, size=0.42\linewidth, connect spies}]
    \node[anchor=south west, inner sep=0] (img) at (0,0)
      {\IfFileExists{#1}%
        {\includegraphics[width=0.79\linewidth]{#1}}%
        {\includegraphics[width=0.79\linewidth]{example-image}}};
    \spy [dashed, red, line width=0.6pt] on (#2,#3)
      in node at (2.11,1.21);
    \useasboundingbox (0,0) rectangle (\linewidth,0 |- img.north);
  \end{tikzpicture}%
}

\begin{figure*}[t]
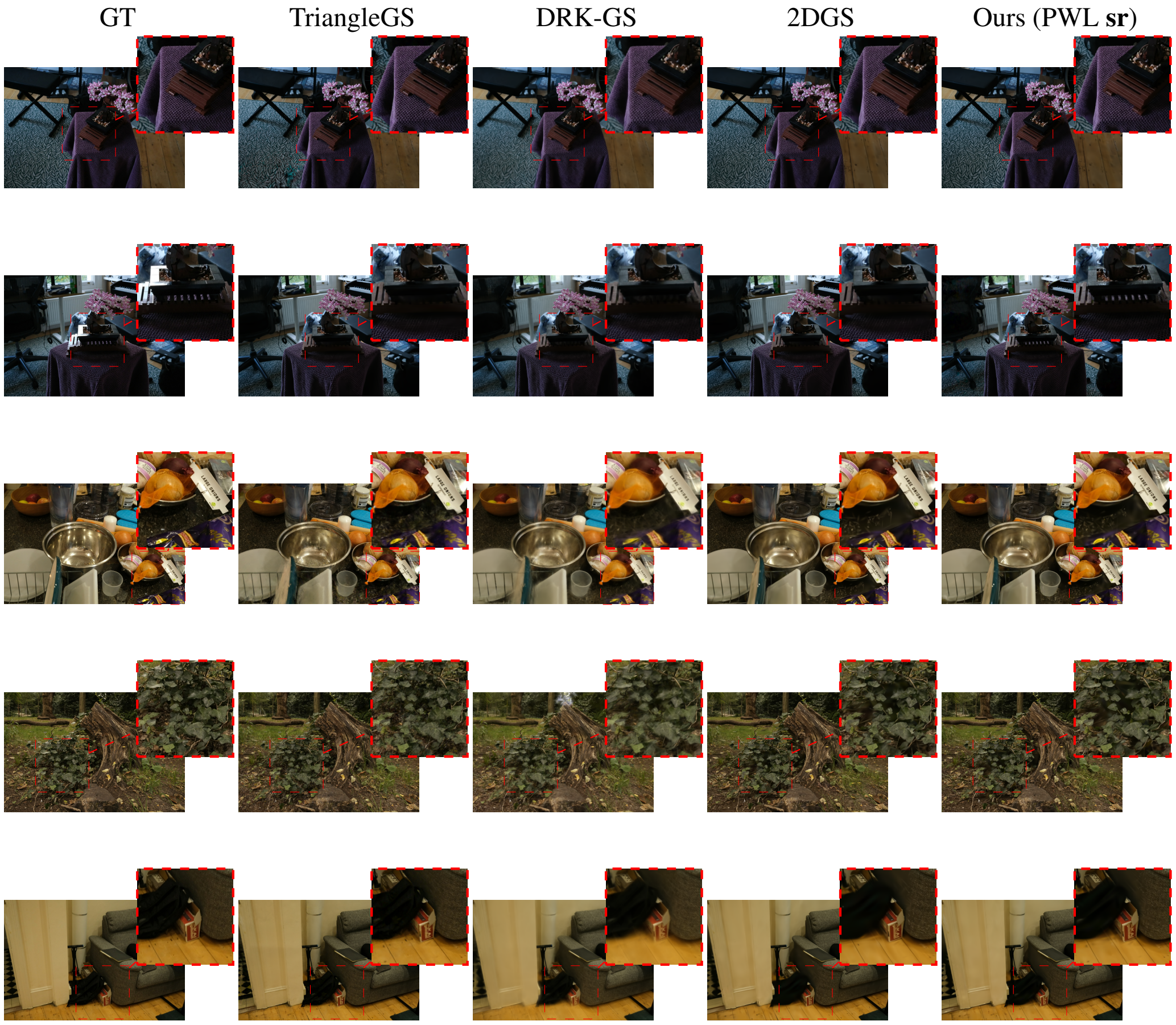

\centering
\setlength{\tabcolsep}{1pt}
\renewcommand{\arraystretch}{0.3}
\newcommand{\imgw}{0.19\textwidth}

\begin{tabular}{ccccc}
  \small GT & \small TriangleGS & \small DRK-GS & \small 2DGS & \small Ours (PWL \textbf{sr}) \\[2pt]

  \begin{minipage}{\imgw}\qualimg{\imgdir/\mGT_\frameA}{0.99}{0.64}{1.8}\end{minipage} &
  \begin{minipage}{\imgw}\qualimg{\imgdir/\mTriangle_\frameA}{0.99}{0.64}{1.8}\end{minipage} &
  \begin{minipage}{\imgw}\qualimg{\imgdir/\mDRK_\frameA}{0.99}{0.64}{1.8}\end{minipage} &
  \begin{minipage}{\imgw}\qualimg{\imgdir/\mTwoDGS_\frameA}{0.99}{0.64}{1.8}\end{minipage} &
  \begin{minipage}{\imgw}\qualimg{\imgdir/\mOurs_\frameA}{0.99}{0.64}{1.8}\end{minipage} \\[40pt]

  \begin{minipage}{\imgw}\qualimg{\imgdir/\mGT_\frameB}{1.09}{0.66}{1.8}\end{minipage} &
  \begin{minipage}{\imgw}\qualimg{\imgdir/\mTriangle_\frameB}{1.09}{0.66}{1.8}\end{minipage} &
  \begin{minipage}{\imgw}\qualimg{\imgdir/\mDRK_\frameB}{1.09}{0.66}{1.8}\end{minipage} &
  \begin{minipage}{\imgw}\qualimg{\imgdir/\mTwoDGS_\frameB}{1.09}{0.66}{1.8}\end{minipage} &
  \begin{minipage}{\imgw}\qualimg{\imgdir/\mOurs_\frameB}{1.09}{0.66}{1.8}\end{minipage} \\[40pt]

  \begin{minipage}{\imgw}\qualimg{\imgdir/\mGT_\frameC}{1.80}{0.31}{1.8}\end{minipage} &
  \begin{minipage}{\imgw}\qualimg{\imgdir/\mTriangle_\frameC}{1.80}{0.31}{1.8}\end{minipage} &
  \begin{minipage}{\imgw}\qualimg{\imgdir/\mDRK_\frameC}{1.80}{0.31}{1.8}\end{minipage} &
  \begin{minipage}{\imgw}\qualimg{\imgdir/\mTwoDGS_\frameC}{1.80}{0.31}{1.8}\end{minipage} &
  \begin{minipage}{\imgw}\qualimg{\imgdir/\mOurs_\frameC}{1.80}{0.31}{1.8}\end{minipage} \\[40pt]

  \begin{minipage}{\imgw}\qualimg{\imgdir/\mGT_\frameD}{0.68}{0.55}{1.8}\end{minipage} &
  \begin{minipage}{\imgw}\qualimg{\imgdir/\mTriangle_\frameD}{0.68}{0.55}{1.8}\end{minipage} &
  \begin{minipage}{\imgw}\qualimg{\imgdir/\mDRK_\frameD}{0.68}{0.55}{1.8}\end{minipage} &
  \begin{minipage}{\imgw}\qualimg{\imgdir/\mTwoDGS_\frameD}{0.68}{0.55}{1.8}\end{minipage} &
  \begin{minipage}{\imgw}\qualimg{\imgdir/\mOurs_\frameD}{0.68}{0.55}{1.8}\end{minipage} \\[40pt]

  \begin{minipage}{\imgw}\qualimg{\imgdir/\mGT_\frameE}{1.15}{0.33}{1.8}\end{minipage} &
  \begin{minipage}{\imgw}\qualimg{\imgdir/\mTriangle_\frameE}{1.15}{0.33}{1.8}\end{minipage} &
  \begin{minipage}{\imgw}\qualimg{\imgdir/\mDRK_\frameE}{1.15}{0.33}{1.8}\end{minipage} &
  \begin{minipage}{\imgw}\qualimg{\imgdir/\mTwoDGS_\frameE}{1.15}{0.33}{1.8}\end{minipage} &
  \begin{minipage}{\imgw}\qualimg{\imgdir/\mOurs_\frameE}{1.15}{0.33}{1.8}\end{minipage} \\
\end{tabular}

\caption{\textbf{Qualitative comparison on MipNeRF 360 ($\times 4$).} Red dashed boxes indicate the zoomed regions.}
\label{fig:qualitative}
\end{figure*}

\section{Pchip Backward Pass Derivation}
\label{app:pchip_backward}
We derive the gradients of the Pchip deformation function with respect to the learnable parameters $a$ and $b$, where $c_1 = 2\sigma(a) - 1$ and $d_1 = 2\sigma(b) - 1$. Given the upstream gradient $\frac{\partial \mathcal{L}}{\partial D}$, the chain rule gives
\begin{equation}
\frac{\partial \mathcal{L}}{\partial a}
= \frac{\partial \mathcal{L}}{\partial D}\,\frac{\partial D}{\partial c_1}\,\frac{\partial c_1}{\partial a}, \qquad
\frac{\partial \mathcal{L}}{\partial b}
= \frac{\partial \mathcal{L}}{\partial D}\,\frac{\partial D}{\partial d_1}\,\frac{\partial d_1}{\partial b}.
\end{equation}
Below we derive $\frac{\partial D}{\partial t}$, $\frac{\partial D}{\partial c_1}$, and $\frac{\partial D}{\partial d_1}$. The final sigmoid chain-rule factors are $\frac{\partial c_1}{\partial a} = 2\sigma(a)(1-\sigma(a))$ and $\frac{\partial d_1}{\partial b} = 2\sigma(b)(1-\sigma(b))$.

\paragraph{Notation.}
The three control points are $\{(-1,-1),\,(c_1,d_1),\,(1,1)\}$. The interval lengths, slopes, and local coordinates on segment$[c_i, c_{i+1}]$ are
\[
h_0 = c_1 + 1, \quad h_1 = 1 - c_1, \quad h_0 + h_1 = 2,
\]
\[
k_0 = \frac{d_1 + 1}{h_0}, \quad k_1 = \frac{1 - d_1}{h_1}, \qquad o = \frac{t - c_i}{h_i}.
\]
The cubic Hermite interpolation is
\begin{equation}
D(t) = H_0(o)\,d_i + H_1(o)\,h_i\,m_i + H_2(o)\,d_{i+1} + H_3(o)\,h_i\,m_{i+1},\quad t \in [c_i, c_{i+1}]
\end{equation}
with basis functions $H_0 = 2o^3 - 3o^2 + 1$, $H_1 = o^3 - 2o^2 + o$, $H_2 = -2o^3 + 3o^2$, $H_3 = o^3 - o^2$, and their derivatives
\begin{align}
H_0' &= 6o^2 - 6o, & H_1' &= 3o^2 - 4o + 1, \nonumber\\
H_2' &= -6o^2 + 6o, & H_3' &= 3o^2 - 2o.
\end{align}
Weights for the interior derivative: $w_1 = 2h_1 + h_0$, $w_2 = h_1 + 2h_0$, so $w_1 + w_2 = 3(h_0 + h_1) = 6$.

\paragraph{Step 1: Gradient with respect to $t$.}
Since $\partial o / \partial t = 1/h_i$:
\begin{equation}
\frac{\partial D}{\partial t} = \frac{1}{h_i}\,\frac{\partial D}{\partial o}, \qquad
\frac{\partial D}{\partial o} = H_0'\,d_i + H_1'\,h_i\,m_i + H_2'\,d_{i+1} + H_3'\,h_i\,m_{i+1}.
\end{equation}

\paragraph{Step 2: Gradients of $k_0$ and $k_1$.}
From $k_0 = (d_1+1)/h_0$ and $k_1 = (1-d_1)/h_1$, using $\partial h_0/\partial c_1 = 1$ and $\partial h_1/\partial c_1 = -1$:
\begin{alignat}{2}
\frac{\partial k_0}{\partial c_1} &= -\frac{k_0}{h_0}, &\qquad \frac{\partial k_0}{\partial d_1} &= \frac{1}{h_0}, \nonumber\\[4pt]
\frac{\partial k_1}{\partial c_1} &= \frac{k_1}{h_1}, &\qquad \frac{\partial k_1}{\partial d_1} &= -\frac{1}{h_1}.
\end{alignat}

\paragraph{Step 3: Gradients of $m_0$ and $m_2$ (endpoint derivatives).}

The unclamped values are
\[
\tilde{m}_0 = \frac{(2h_0 + h_1)\,k_0 - h_0\,k_1}{2}, \qquad
\tilde{m}_2 = \frac{(2h_1 + h_0)\,k_1 - h_1\,k_0}{2}.
\]

\noindent\textbf{Case $\tilde{m}_0 \geq 0$} (clamping inactive, $m_0 = \tilde{m}_0$). Let $N_0 = (2h_0 + h_1)k_0 - h_0 k_1$.

\emph{Gradient w.r.t.\ $c_1$:}
\begin{align}
\frac{\partial N_0}{\partial c_1}
&= \underbrace{\frac{\partial(2h_0+h_1)}{\partial c_1}}_{= 1}\,k_0
  + (2h_0+h_1)\!\left(-\frac{k_0}{h_0}\right)
  - \underbrace{\frac{\partial h_0}{\partial c_1}}_{= 1}\,k_1
  - h_0\,\frac{k_1}{h_1} \nonumber\\
&= k_0 - \frac{(2h_0+h_1)}{h_0}\,k_0 - k_1 - \frac{h_0}{h_1}\,k_1 \nonumber\\
&= k_0\!\left(1 - 2 - \frac{h_1}{h_0}\right) - k_1\!\left(1 + \frac{h_0}{h_1}\right) \nonumber\\
&= -\frac{h_0 + h_1}{h_0}\,k_0 - \frac{h_0 + h_1}{h_1}\,k_1
= -2\!\left(\frac{k_0}{h_0} + \frac{k_1}{h_1}\right).
\end{align}
Therefore
\begin{equation}\label{eq:dm0_dc1}
\frac{\partial m_0}{\partial c_1} = \frac{1}{2}\,\frac{\partial N_0}{\partial c_1} = -\frac{k_0}{h_0} - \frac{k_1}{h_1}.
\end{equation}

\emph{Gradient w.r.t.\ $d_1$:}
\begin{align}
\frac{\partial N_0}{\partial d_1}
&= (2h_0+h_1)\,\frac{1}{h_0} + h_0\,\frac{1}{h_1}
= 2 + \frac{h_1}{h_0} + \frac{h_0}{h_1}.
\end{align}
Therefore
\begin{equation}\label{eq:dm0_dd1}
\frac{\partial m_0}{\partial d_1} = 1 + \frac{h_0^2 + h_1^2}{2\,h_0\,h_1}.
\end{equation}

\noindent\textbf{Case $\tilde{m}_0 < 0$} (clamping active, $m_0 = 0$): $\frac{\partial m_0}{\partial c_1} = \frac{\partial m_0}{\partial d_1} = 0$.

\medskip

By the symmetry of $\tilde{m}_2$, when clamping is inactive ($\tilde{m}_2 \geq 0$):
\begin{equation}\label{eq:dm2}
\frac{\partial m_2}{\partial c_1} = \frac{k_0}{h_0} + \frac{k_1}{h_1}
= -\frac{\partial m_0}{\partial c_1}, \qquad
\frac{\partial m_2}{\partial d_1} = -1 - \frac{h_0^2 + h_1^2}{2\,h_0\,h_1}
= -\frac{\partial m_0}{\partial d_1}.
\end{equation}

\paragraph{Step 4: Gradients of $m_1$ (interior derivative).}

\noindent\textbf{Case $k_0 k_1 \leq 0$:} $m_1 = 0$, all gradients vanish.

\noindent\textbf{Case $k_0 k_1 > 0$:} Define $\tau = w_1/k_0 + w_2/k_1$, so $m_1 = (w_1+w_2)/\tau = 6/\tau$. Since $w_1 + w_2 = 6$ is constant:
\begin{equation}
\frac{\partial m_1}{\partial \theta}
= -\frac{6}{\tau^2}\,\frac{\partial \tau}{\partial \theta}
= -\frac{m_1^2}{6}\,\frac{\partial \tau}{\partial \theta}, \qquad \theta \in \{c_1, d_1\}.
\end{equation}

\emph{Gradient of $\tau$ w.r.t.\ $c_1$:}
Using $\partial w_1/\partial c_1 = -1$, $\partial w_2/\partial c_1 = 1$, and $\frac{\partial(1/k)}{\partial c_1} = -\frac{1}{k^2}\frac{\partial k}{\partial c_1}$:
\begin{align}
\frac{\partial \tau}{\partial c_1}
&= \frac{-1}{k_0} + w_1\!\left(\frac{1}{h_0\,k_0}\right)
   + \frac{1}{k_1} + w_2\!\left(-\frac{1}{h_1\,k_1}\right) \nonumber\\
&= \frac{w_1 - h_0}{h_0\,k_0} + \frac{h_1 - w_2}{h_1\,k_1}
 = \frac{2h_1}{h_0\,k_0} - \frac{2h_0}{h_1\,k_1},
\end{align}
where we used $w_1 - h_0 = 2h_1$ and $w_2 - h_1 = 2h_0$.

\emph{Gradient of $\tau$ w.r.t.\ $d_1$:}
Since $w_1, w_2$ do not depend on $d_1$:
\begin{equation}
\frac{\partial \tau}{\partial d_1}
= w_1\!\left(-\frac{1}{k_0^2}\cdot\frac{1}{h_0}\right) + w_2\!\left(-\frac{1}{k_1^2}\cdot\left(-\frac{1}{h_1}\right)\right)
= -\frac{w_1}{h_0\,k_0^2} + \frac{w_2}{h_1\,k_1^2}.
\end{equation}

Therefore:
\begin{equation}
\frac{\partial m_1}{\partial c_1} = -\frac{m_1^2}{6}\!\left(\frac{2h_1}{h_0\,k_0} - \frac{2h_0}{h_1\,k_1}\right), \qquad
\frac{\partial m_1}{\partial d_1} = -\frac{m_1^2}{6}\!\left(-\frac{w_1}{h_0\,k_0^2} + \frac{w_2}{h_1\,k_1^2}\right).
\end{equation}

\paragraph{Step 5: Per-segment assembly.}

On each segment, $D$ depends on $c_1$ and $d_1$ through (i)~the local coordinate $o$, (ii)~the interval length $h_i$ (which scales the derivative terms), (iii)~the values $d_i$, $d_{i+1}$, and (iv)~the derivatives $m_i$, $m_{i+1}$.

\medskip
\noindent\textbf{Segment $i = 0$: $t \in [-1, c_1]$}, with $d_0 = -1$, $d_1 = d_1$, $o = (t+1)/h_0$.

\emph{Dependence of $o$ on $c_1$:} Since $h_0 = c_1 + 1$,
\[
\frac{\partial o}{\partial c_1} = -\frac{t+1}{h_0^2} = -\frac{o}{h_0}.
\]

The full expression $D = H_0(-1) + H_1 h_0 m_0 + H_2 d_1 + H_3 h_0 m_1$ depends on $c_1$ through $o$, $h_0$, $m_0$, and $m_1$:
\begin{align}
\frac{\partial D}{\partial c_1}\bigg|_{i=0}
&= \frac{\partial D}{\partial o}\cdot\frac{\partial o}{\partial c_1}
  + \frac{\partial}{\partial c_1}\!\Big[H_1\,h_0\,m_0 + H_3\,h_0\,m_1\Big] \nonumber\\
&= \frac{\partial D}{\partial o}\!\left(-\frac{o}{h_0}\right)
  + (H_1\,m_0 + H_3\,m_1)
  + H_1\,h_0\,\frac{\partial m_0}{\partial c_1}
  + H_3\,h_0\,\frac{\partial m_1}{\partial c_1}.
\end{align}
The term $(H_1 m_0 + H_3 m_1)$ arises from $\frac{\partial h_0}{\partial c_1} = 1$.

For $d_1$, noting that $d_{i+1} = d_1$ appears via $H_2$, and $h_0$, $o$ do not depend on $d_1$:
\begin{equation}
\frac{\partial D}{\partial d_1}\bigg|_{i=0}
= H_2
  + H_1\,h_0\,\frac{\partial m_0}{\partial d_1}
  + H_3\,h_0\,\frac{\partial m_1}{\partial d_1}.
\end{equation}

\medskip
\noindent\textbf{Segment $i = 1$: $t \in [c_1, 1]$}, with $d_1 = d_1$, $d_2 = 1$, $o = (t - c_1)/h_1$.

\emph{Dependence of $o$ on $c_1$:} Since $h_1 = 1 - c_1$,
\[
\frac{\partial o}{\partial c_1}
= \frac{-h_1 - (t - c_1)(-1)}{h_1^2}
= \frac{-h_1 + (t - c_1)}{h_1^2}
= \frac{t - c_1 - h_1}{h_1^2}
= \frac{t - 1}{h_1^2}.
\]

The full expression $D = H_0 d_1 + H_1 h_1 m_1 + H_2(1) + H_3 h_1 m_2$ depends on $c_1$ through $o$, $h_1$, $m_1$, and $m_2$:
\begin{align}
\frac{\partial D}{\partial c_1}\bigg|_{i=1}
&= \frac{\partial D}{\partial o}\cdot\frac{t-1}{h_1^2}
  + \frac{\partial}{\partial c_1}\!\Big[H_1\,h_1\,m_1 + H_3\,h_1\,m_2\Big] \nonumber\\
&= \frac{\partial D}{\partial o}\!\left(\frac{t-1}{h_1^2}\right)
  - (H_1\,m_1 + H_3\,m_2)
  + H_1\,h_1\,\frac{\partial m_1}{\partial c_1}
  + H_3\,h_1\,\frac{\partial m_2}{\partial c_1}.
\end{align}
The term $-(H_1 m_1 + H_3 m_2)$ arises from $\frac{\partial h_1}{\partial c_1} = -1$.

For $d_1$, noting that $d_i = d_1$ appears via $H_0$:
\begin{equation}
\frac{\partial D}{\partial d_1}\bigg|_{i=1}
= H_0
  + H_1\,h_1\,\frac{\partial m_1}{\partial d_1}
  + H_3\,h_1\,\frac{\partial m_2}{\partial d_1}.
\end{equation}


\paragraph{Step 6: Chain rule to learnable parameters.}

\begin{align}
\frac{\partial \mathcal{L}}{\partial a}
&= \frac{\partial \mathcal{L}}{\partial D}\,\frac{\partial D}{\partial c_1}\,\cdot\, 2\sigma(a)\bigl(1-\sigma(a)\bigr), \\
\frac{\partial \mathcal{L}}{\partial b}
&= \frac{\partial \mathcal{L}}{\partial D}\,\frac{\partial D}{\partial d_1}\,\cdot\, 2\sigma(b)\bigl(1-\sigma(b)\bigr).
\end{align}
\newpage

\end{document}